\documentstyle[12pt,fullpage]{article}
\input psfig
\newcommand{\bbox}{\vrule height7pt width4pt depth1pt}
\newcommand{\metapredfont}{\sc}
\newcommand{\Poss}{\mbox{\metapredfont Poss}}
\newcommand{\Know}{\mbox{\metapredfont Know}}
\newcommand{\ObsIn}{\mbox{\metapredfont Oi}}
\newcommand{\pr}{\mbox{\it p}}

\newcommand{\bel}{\mbox{\metapredfont Bel}}
\newcommand{\K}{\mbox{$K$}}
\newcommand{\sump}{\mbox{sum$_{\it p}$}}
\newcommand{\suml}{\mbox{sum$_{\ell}$}}

\newcommand{\lhood}{\mbox{$\ell$}}

\newcommand{\norm}{\mbox{\it Normal}}
\newcommand{\dofn}{\mbox{\it do}}
\newcommand{\dofnsub}{\mbox{\scriptsize \it do}}
\newcommand{\Do}{\mbox{\it Do}}

\newcommand{\actionfont}{\small\sf}

\newcommand{\pickup}{\mbox{\actionfont pickup}}
\newcommand{\readcombi}{\mbox{\actionfont read-combi}}
\newcommand{\readPaper}{\mbox{\actionfont read}}
\newcommand{\putdown}{\mbox{\actionfont putdown}}
\newcommand{\myput}{\mbox{\actionfont put}}
\newcommand{\walk}{\mbox{\actionfont walk}}

\newcommand{\repair}{\mbox{\actionfont repair}}

\newcommand{\explode}{\mbox{\actionfont explode}}

\newcommand{\exactAdvance}{\mbox{\actionfont exact-advance}}
\newcommand{\noisyAdvance}{\mbox{\actionfont noisy-advance}}
\newcommand{\theAdvance}{\mbox{\actionfont advance}}

\newcommand{\noisyPos}{\mbox{\actionfont noisy-sense-position}}
\newcommand{\thePos}{\mbox{\actionfont sense-position}}

\newcommand{\thesensef}{\mbox{\actionfont sense-f}}
\newcommand{\noisysensef}{\mbox{\actionfont noisy-sense-f}}
\newcommand{\thef}{\mbox{\actionfont change-f}}
\newcommand{\thefsub}{\mbox{\actionfont \scriptsize change-f}}
\newcommand{\noisyf}{\mbox{\actionfont noisy-change-f}}

\newcommand{\drop}{\mbox{\actionfont drop}}
\newcommand{\dropbreak}{\mbox{\actionfont drop-break}}
\newcommand{\dropnobreak}{\mbox{\actionfont drop-not-break}}

\newcommand{\holding}{\mbox{\it Holding\/}}
\newcommand{\fragile}{\mbox{\it Fragile\/}}

\newcommand{\broken}{\mbox{\it Broken\/}}
\newcommand{\book}{\mbox{\it Book\/}}
\newcommand{\light}{\mbox{\it Light-On\/}}

\newcommand{\nextto}{\mbox{\it NextTo\/}}
\newcommand{\iscarrying}{\mbox{\it IsCarrying\/}}
\newcommand{\position}{\mbox{\it position\/}}
\newcommand{\combi}{\mbox{\it combi\/}}
\newcommand{\heavy}{\mbox{\it Heavy\/}}
\newcommand{\colour}{\mbox{\it Colour\/}}
\newcommand{\on}{\mbox{\it On\/}}
\newcommand{\prev}{\mbox{\it prev\/}}
\newcommand{\slippery}{\mbox{\it slippery\/}}
\newcommand{\snow}{s_{\it now}}
\newcommand{\anow}{a_{\it now}}

\newcommand{\succes}[1]{{#1}^+}

\newcommand{\eqdef}{\;\stackrel{\rm def}{=}\;}

\newcommand{\true}{\mbox{\sc true}}

\newtheorem{EXAMPLE}{Example}[section]
\newenvironment{example}
{\begin{EXAMPLE} \hspace{-.75em} {\bf :} \rm}{\bbox\end{EXAMPLE}}

\newtheorem{DEFINITION}{Definition}
\newenvironment{defn}
{\begin{DEFINITION} \hspace{-.75em} {\bf :} \rm}{\bbox\end{DEFINITION}}

\begin{document}

\begin{titlepage}
\thispagestyle{empty}
\title{Reasoning about Noisy Sensors and Effectors in the Situation
  Calculus\thanks{The work of Fahiem Bacchus and Hector Levesque was
    supported in part by the Canadian government through their NSERC
    and IRIS programs. The work of Joseph Halpern supported in part by
    the Air Force Office of Scientific Research (AFSC), under
    Contracts F49620-91-C-0080
and F94620-96-1-0323 and NSF grant IRI-9625901.  Much of this work was
done while Halpern was at the IBM Almaden Research Center.
    A preliminary version of
    this paper appears in {\em Proceedings of IJCAI '95},
    pp.~1933--1940.}}

\author{ \renewcommand{\arraystretch}{.70}
\begin{tabular}[t]{c}
Fahiem Bacchus\\
   {\small Dept. Computer Science}\\
   {\small University of Waterloo}\\
   {\small Waterloo, Ontario}\\
   {\small Canada, N2L 3G1}\\
   {\small fbacchus@logos.uwaterloo.ca}
\end{tabular}
\renewcommand{\arraystretch}{.70}
\begin{tabular}[t]{c}
   Joseph Y.\ Halpern\\
   {\small Dept. Computer Science}\\
   {\small Cornell University}\\
   {\small Ithaca, NY 14850}\\
   {\small halpern@cs.cornell.edu}\\
\end{tabular}
\renewcommand{\arraystretch}{.70}
\begin{tabular}[t]{c}
   Hector J.\ Levesque\\
   {\small Dept. Computer Science}\\
   {\small University of Toronto}\\
   {\small Toronto, Ontario}\\
   {\small Canada, M5S 3H5}\\
   {\small hector@cs.toronto.edu}
\end{tabular}
}

\maketitle
\thispagestyle{empty}

\begin{abstract}
  Agents interacting with an incompletely known
world need to
  be able to reason about the effects of their actions, and to gain
  further information about that world they need to use sensors of
  some sort. Unfortunately, both the effects of actions and the
  information returned from sensors are subject to error.
  To cope with such uncertainties, the agent can maintain probabilistic
  beliefs about the state of the world. With probabilistic beliefs the
  agent will be able to quantify the likelihood of the various
  outcomes of its actions
and
is better able to utilize the information gathered
  from its error-prone actions and sensors.
  In this paper, we present a model in which we can reason about an
  agent's probabilistic degrees of belief and the manner in which
  these beliefs change as various actions are executed. We build on a
  general logical theory of action developed by Reiter and others,
  formalized in the situation calculus.  We propose a simple
  axiomatization that captures an agent's state of belief and the
  manner in which these beliefs change when actions are executed. Our
  model displays a number of intuitively reasonable properties.
\end{abstract}
\end{titlepage}

\section{Introduction}

An intelligent agent interacting with a dynamic and incompletely known
world faces two special sorts of reasoning problems. First, because
the world is {\em dynamic}, it will need to reason about change: how
its actions and the actions of others affect the state of the world.
For example, an agent will need to reason that if a fragile object is
dropped then it will probably break, and regardless of what else
happens, the object will remain broken until it is repaired. Second,
because the world is {\em incompletely known}, the agent will need to
make do with partial descriptions of the state of the world. As a
result, the agent will often need to augment what it knows by
performing perceptual actions, using sensors of some sort.
Unfortunately, both the effectors that the agent uses to modify the
world and the sensors it uses to sense the world will in practice be
subject to uncertainty, i.e., they will be noisy. For example, a
robotic agent may not know initially how far away it is from the
nearest wall, but may have a sensor that it can use to obtain
information about this distance. Because the sensor is noisy, a
reading of, say, ``3.1 meters'' does not guarantee that the agent is
actually 3.1 meters from the wall, although it should serve to
increase the agent's degree of belief in that fact. It may need to
read this sensor (or additional sensors) a number of times to get a
sufficiently reliable measurement. Similarly, if it attempts to move
0.7 meters towards the wall, it may well end up moving 0.8 meters due
to inaccuracies in its effectors. Nevertheless, its degree of belief
that it is closer to the wall should increase. In this paper, we
propose a representational formalism to capture the reasoning required
to keep the agent's beliefs about the world synchronized with the
effects of the various actions it performs. Without such a
synchronization the agent would be unable to modify its environment in
any purposeful manner.

Somewhat surprisingly, although the importance of dealing with dynamic
and incompletely known worlds has long been argued within AI, very few
adequate representation formalisms have emerged. We can classify
existing ones into two broad camps. On the one hand, we have
probabilistic formalisms such as Bayesian nets
\cite{Pearl:Probabilistic.Reasoning.Intelligent.Systems} for dealing
with uncertainty in general, and the uncertainty that would arise from
noisy sensors in particular. However, with the exceptions noted below,
these probabilistic formalisms have not attempted to incorporate a
general model of action, i.e., representing what does and does not
change as the result of performing an action. In addition, while it is
possible to express in these formalisms probabilistic dependencies
among variables, which are in essence atomic propositions, it is not
easy to deal with many other forms of incomplete knowledge. For
example, it is difficult to say that one of two conditions holds, or
that all objects of a certain type have a certain property when it is
not known what those objects are. Logical formalisms, on the other
hand, with features like disjunction and quantification, are well
suited for expressing incomplete knowledge of this type. Moreover,
logical formalisms like dynamic logics, process logics, or the
situation calculus, allow us to reason about the prerequisites and
effects of actions.

These observations suggest that it would be useful to have a formalism
that allows us to combine actions, knowledge, (probabilistic) beliefs,
and first-order reasoning. There has been work that puts some of these
components together. For example, the frameworks of Bacchus
\cite{Bacchus} and Halpern \cite{Hal4} combines first-order reasoning
and probability, but this work does not explicitly address reasoning
about actions.  The framework of Halpern and Tuttle \cite{HT} has
probability, knowledge, time, and actions, but it is not a first-order
formalism. Finally, the action network formalism of Darwiche and
Goldszmidt \cite{DG:UAI94} extends Bayesian nets to allow
probabilistic reasoning about action and observation sequences and
their effects, but it also is not a first-order formalism, nor can it
easily deal with other forms of incomplete knowledge like disjunction.

In this paper, we propose a framework that combines first-order
reasoning, with reasoning about knowledge, probability, and actions.
For reasoning about actions we start with a variant of the situation
calculus \cite{mccarthy-hayes:sitcalc} that incorporates the solution
to the frame problem proposed by Reiter \cite{reiter:frame}. Building
on this and on the work of Scherl and Levesque \cite{scherl-levesque},
who incorporated knowledge and perceptual actions into the situation
calculus, we show how probabilities can be added and how the effects
of actions on an agent's probabilistic beliefs can be modeled.

We base our work on the situation calculus as it has proved to be a very 
convenient formalism for modeling actions, their prerequisites, and their 
effects, and for modeling incomplete knowledge of the state of the world. 
Of course, our approach inherits these features. In addition, it allows us 
to build on previous work done in this framework and to take advantage of 
parallel developments. For example, although Reiter's solution to the frame 
problem is limited in a number of ways, it has been extended to handle 
aspects of the ramification problem \cite{lin-reiter:ram}, agent ability 
\cite{lesperance:ability}, continuous time \cite{pinto:phd}, and perceptual 
actions \cite{scherl-levesque}. Another extension of the theory to deal 
with {\em complex actions\/} (sequence, iterations, recursive procedures, 
non-determinism, etc.), briefly described in 
Section~\ref{sec:nondeterminism}, has led to a novel logic programming 
language called {\sc golog}. {\sc Golog} has proved to be useful for 
describing high-level robot and softbot control \cite{gologpaper}. An 
implementation of {\sc golog} exists and a number of programs (including a 
banking agent softbot application \cite{Ruman97}) currently run in 
simulation mode. A mail delivery application written in {\sc golog} also 
runs on two different robot platforms \cite{Jenkin97}. We expect that the 
extensions we make here can be incorporated into this framework as well.

Independently of the situation calculus, however, our formalism
demonstrates an interesting interaction between actions and differing
mechanisms for updating probabilistic beliefs. And as we hope to make
clear {}from our presentation, much of what we do here could be
carried out in other logical frameworks.

The rest of the paper is organized as follows. In the next section, we
briefly review the theory of action in terms of which our account is
formulated: the situation calculus and the solution to the frame
problem proposed by Reiter. In Section~\ref{sec:nondeterminism}, we
examine how non-deterministic actions can be handled within this
framework. In Section~\ref{sec:epistemic}, we augment the framework
by adding the notion of the agent's epistemic state. In
Section~\ref{sec:beliefs} we show how probabilities can be added, and
present a simple formalization within the situation calculus of the
degree of belief an agent has in propositions expressed as logical
formulas. This allows us to formalize in more quantitative terms the
changes in belief that arise when dealing with noisy sensors and
effectors. Examples of the formalism at work are presented in
Section~\ref{sec:examples}, and some conclusions are drawn in
Section~\ref{sec:conclusion}.

\section{A Theory of Action}\label{sec:sitcalc}
Our account of sensors is formulated as a logical theory in an
extended version of the situation calculus. The situation calculus
\cite{mccarthy-hayes:sitcalc} is a many-sorted dialect of the
predicate calculus (with some second-order features), containing sorts
for (among other things) {\em situations}, which are like the possible
worlds of modal logic, for primitive deterministic {\em actions}, and,
since we will be dealing with probabilities, for {\em real numbers}.

The situation calculus is specifically designed for representing
dynamically changing worlds. All changes to the world are the result
of named {\em actions}. A possible world history, which is simply a
sequence of actions, is represented by a first-order term called a
{\em situation}. The constant $S_0$ is used to denote the {\em initial
  situation}, namely that situation in which no actions have yet
occurred. There is a distinguished binary function symbol $\dofn$;
$\dofn(\alpha,s)$ denotes the successor situation to $s$ resulting
{}from performing the action $\alpha$. Actions may be parameterized. For
example, $\myput(x,y)$ might stand for the action of putting object
$x$ on object $y$, in which case $\dofn(\myput(A,B),s)$ denotes that
situation resulting from placing $A$ on $B$ when in situation $s$.
Notice that in the situation calculus, actions are denoted by first
order terms, and situations (world histories) are also first order
terms. For example, $\dofn(\putdown(A)$, $\dofn(\walk(L)$,
$\dofn(\pickup(A),S_0)))$ is a situation denoting the world history
consisting of the sequence of actions $[\pickup(A)$, $\walk(L)$,
$\putdown(A)]$.

Relations whose truth values vary from situation to situation, called
{\em relational fluents\/}, are denoted by predicate symbols taking a
situation term as their last argument. For example,
$\iscarrying(\mbox{\it robot},p,s)$, meaning that the {\it robot\/} is
carrying package $p$ in situation $s$, is a relational fluent.
Functions whose denotations vary from situation to situation are
called {\em functional fluents\/}. They are denoted by function
symbols with an extra argument taking a situation term, as in
$\position(\mbox{\it robot},s)$, i.e., the robot's location in
situation $s$. Finally, we use the distinguished predicate $\Poss(a,s)$ to 
state that action $a$ is possible to execute in situation $s$.

\subsection{Actions}
Actions have {\em preconditions}---necessary and sufficient conditions
that characterize when the action is physically possible. For example,
in a blocks world, we might have:%
\footnote{In formulas, free variables are considered to be
  universally quantified.
  This convention will be
  followed throughout the paper.}
\[
\begin{array}{l}
  \Poss(\pickup(x),s) \\
  \qquad \mbox{}\equiv
  \lnot\exists z.\,\holding(z,s) \land \nextto(x,s) \land \lnot \heavy(x),
\end{array}
\]
which says that an object $x$ can be picked up if and only if the
agent is not currently holding something, it is next to $x$, and $x$ is
not heavy.

\begin{defn}[Precondition Axioms]
  \label{def:preaxioms}
  {\em Precondition axioms\/} are axioms of the form
  \[\Poss(a,s) \equiv \phi(a,s). \]
  Here $\phi$ is a formula that specifies the necessary and sufficient
  properties of the situation $s$ and the action $a$, under which $a$
  can be executed in $s$.
\end{defn}

The manner in which the actions modify the world are specified by
their {\it effect axioms}. These describe the effects of a given
action on the fluents. For example, a robot dropping a fragile object
causes it to be broken:
\begin{equation}
  \label{broken1}
  \Poss(\drop(r,x),s) \land \fragile(x,s) \supset
  \broken(x,\dofn(\drop(r,x),s)).
\end{equation}

Exploding a bomb next to an object causes it to be broken:
\begin{equation}
  \label{broken2}
  \Poss(\explode(b),s) \land \nextto(b,x,s) \supset
  \broken(x, \dofn(\explode(b),s)).
\end{equation}

A robot repairing an object causes it to be not broken:
\begin{equation}
  \label{broken3}
  \Poss(\repair(r,x),s) \supset \lnot \broken(x,\dofn(\repair(r,x),s)).
\end{equation}

\subsection{The Frame Problem}
As first observed by McCarthy and Hayes \cite{mccarthy-hayes:sitcalc},
axiomatizing a dynamic world requires more than just action
precondition and effect axioms. So-called {\em frame axioms\/} are
also necessary. These specify the action {\em invariants\/} of the
domain, namely, those fluents that remain unaffected by a given
action. For example, an object's colour is not affected by a robot
dropping something:
\[
\Poss(\drop(r,x),s) \land \colour(y,s) = c
\supset \colour(y,\dofn(\drop(r,x),s)) = c.
\]
A frame axiom describing how the fluent $\broken$ is unaffected:
\[
\begin{array}{l}
\Poss(\drop(r,x),s) \land \lnot \broken(y,s)
\land y \neq x \lor \lnot \fragile(y,s) \\
\qquad\mbox{} \supset \lnot \broken(y,\dofn(\drop(r,x),s)).
\end{array}
\]

The problem introduced by the need for such frame axioms is that we
can expect a
great many
of them. Relatively few actions will affect
the truth value of a given fluent; all other actions leave the fluent
unchanged. For example, an object's colour is not changed by picking
things up, opening a door, going for a walk, electing a new prime
minister of Canada, etc. This is problematic for the axiomatizer who
must think of all these axioms; it is also problematic for any
automated reasoning process as it must reason efficiently in the
presence of so many frame axioms.

Suppose that the person responsible for axiomatizing an application
domain
has specified all of the causal laws for the world being axiomatized.
More precisely, suppose they have succeeded in writing down {\em
  all\/} the effect axioms, i.e., for each fluent $F$ and each action
$A$ that can cause $F$'s truth value to change,
they have written
axioms of the form
\[
\Poss(A,s) \land R({\vec x},s) \supset (\lnot)F({\vec x},\dofn(A,s)),
\]
where $R$ is a first-order formula specifying the contextual conditions
under which the action $A$ will have its specified effect on $F$.

A reasonable solution to the frame problem would be a systematic
procedure for generating, from these effect axioms, all the frame
axioms. If possible, we would also want a {\em parsimonious\/}
representation for these frame axioms (because in their simplest form,
there are too many of them).

\subsection{A Simple Solution to the Frame Problem}
\label{sec:framesol}
By appealing to earlier ideas of Haas \cite{Haas87}, Schubert
\cite{Schubert90}, and Pednault \cite{Pednault89}, Reiter
\cite{reiter:frame} proposed a simple solution to the frame problem,
which we employ in this work. It is best illustrated with an
example. Suppose that Equations \ref{broken1}, \ref{broken2}, and
\ref{broken3} are all the effect axioms for the fluent $\broken$,
i.e., they describe all the ways that any action can change the truth
value of $\broken$. We can rewrite
(\ref{broken1}) and (\ref{broken2})
in the logically equivalent form:
\[
  \begin{array}{l}
    \Poss(a,s) \land \exists r.\,a = \drop(r,x) \land \fragile(x,s)\\
    \qquad \mbox{} \lor \exists b.\,a = \explode(b) \land \nextto(b,x,s)\\
    \qquad \qquad \supset \broken(x,\dofn(a,s)).
  \end{array}
\]
Similarly, consider the negative effect axiom for $\broken$,
(\ref{broken3}); this can be rewritten as:
\[
  \Poss(a,s) \land \exists r.\,a = \repair(r,x) \supset \lnot
  \broken(x,\dofn(a,s)).
\]

In general, we can assume that the effect axioms for a fluent $F$ have
been written in the forms:
\begin{equation}
  \label{poseffect}
  \Poss(a,s) \land \gamma_{F}^{+}(\vec{x},a,s) \supset F(\vec{x},\dofn(a,s)),
\end{equation}
\begin{equation}
  \label{negeffect}
  \Poss(a,s) \land \gamma_{F}^{-}(\vec{x},a,s) \supset
  \lnot F(\vec{x},\dofn(a,s)).
\end{equation}
Here $\gamma_{F}^{+}(\vec{x},a,s)$ is a formula describing the
conditions under which doing the action $a$ in situation $s$ causes the
fluent $F$ to become true in the successor situation $\dofn(a,s)$;
similarly $\gamma_{F}^{-}(\vec{x},a,s)$ describes the conditions under
which performing $a$ in $s$ causes $F$ to become false in the
successor situation. Reiter's solution to the frame problem rests on a
{\em completeness assumption\/}, which is that the causal axioms
(\ref{poseffect}) and (\ref{negeffect}) characterize all the
conditions under which action $a$ can lead to a fluent $F(\vec{x})$
becoming true (respectively, false) in the successor situation. In
other words, axioms (\ref{poseffect}) and (\ref{negeffect})
describe all the causal laws affecting the truth values of the fluent
$F$. Therefore, if action $a$ is possible and $F(\vec{x})$'s truth
value changes from {\em false\/} to {\em true\/} as a result of doing
$a$, then $\gamma^{+}_{F}(\vec{x},a,s)$ must be {\em true\/} and
similarly for a change from {\em true\/} to {\em false\/}.  Reiter
\cite{reiter:frame} shows how to automatically derive a
{\em successor-state axiom\/} from the causal axioms (\ref{poseffect}) and
(\ref{negeffect}) and the completeness assumption.

\begin{defn}[Successor-State Axioms]
  \label{def:ssaxioms}
  {\em Successor-state axioms\/} are axioms of the following form
  \[
  \begin{array}{l}
    \Poss(a,s) \supset F(\vec{x},\dofn(a,s)) \\
    \qquad \mbox{} \equiv
    \gamma_{F}^{+}(\vec{x},a,s) \lor (F(\vec{x},s) \land \lnot
    \gamma_{F}^{-}(\vec{x},a,s)).
  \end{array}
  \]
  for relational fluents $F$, and of the form
  \[
  \begin{array}{l}
    \Poss(a,s) \supset f(\vec{x},\dofn(a,s)) = z \mbox{} \\
    \qquad \mbox{} \equiv
    \gamma_{f}(z,\vec{x},a,s),
  \end{array}
  \]
  for functional fluents $f$. These axioms characterize the state of a
  fluent $F$ (or the value of a functional fluent $f$) in the
  successor situation given properties of the current state and the
  action executed.
\end{defn}

The successor-state axiom for $F$ is a solution to the frame problem for
that fluent. Notice that this axiom universally quantifies over
actions $a$. In fact,
the ability to universally quantify over actions is one of the keys to
obtaining
a parsimonious solution to the frame problem.

Applying this to our example about breaking things, we obtain the
following successor-state axiom:
\[
\begin{array}{l}
  \Poss(a,s) \supset \broken(x,\dofn(a,s)) \\
  \qquad \mbox{} \equiv
  \exists r.\,a = \drop(r,x) \land \fragile(x,s)\\
  \qquad\qquad \mbox{} \lor \exists b.\,a = \explode(b) \land \nextto(b,x,s)\\
  \qquad\qquad \mbox{} \lor \broken(x,s) \land \lnot \exists r.\,a =
  repair(r,x).
\end{array}
\]

It is important to note that the above solution to the frame problem
presupposes that there are no {\em state constraints}, as for example
the blocks world constraint: $\forall s.\,\on(x,y,s) \supset \lnot
\on(y,x,s)$.  Such constraints can implicitly contain effect axioms
(so-called indirect effects), in which case the above completeness
assumption will not be true. The assumption that there are no state
constraints in the axiomatization of the domain will be
made throughout this paper.%
\footnote{In \cite{lin-reiter:ram}, the approach discussed in this
  section is extended to deal with some forms of state constraints, by
  compiling their effects into the successor-state axioms.}

In general, 
for any application, we use what is called in \cite{lin-reiter:ram}
a {\em basic action theory} consisting of the following axioms:
\begin{itemize}

\item axioms describing the initial situation---what is true
  initially, before any actions have occurred. This is any finite set
  of sentences that mention only
  the situation term $S_0$.

\item action-precondition axioms (Defn.~\ref{def:preaxioms}), one for
  each primitive action

\item successor-state axioms (Defn.~\ref{def:ssaxioms}), one for each
  fluent

\item unique-name axioms for the primitive actions
(saying that primitive actions with distinct names are different)

\item
  a set of domain-independent foundational axioms for the situation
  calculus \cite{lin-reiter:ram}. These are used to state that the
  situations are all and only those reachable from $S_0$ by performing
  a finite sequence of actions.

\end{itemize}

\section{Nondeterministic Actions}
\label{sec:nondeterminism}

In the description of the situation calculus in the previous section,
actions are taken
to be deterministic in the sense that it is assumed that there exists
a {\em unique\/} successor state for any action executed.  To model
noisy sensors and effectors, we first show how nondeterministic
actions can be handled within this framework without giving up the
simple solution to the frame problem outlined above.

Nondeterministic actions are treated as being the actual execution of
one of a range of primitive (deterministic) actions. Nondeterminism
arises from the fact that we do not know exactly which primitive
action was actually executed. Thus, when a nondeterministic action is
executed by the agent, the current situation will have a {\em set\/}
of successor states, one successor state for each possible primitive
action.  Nevertheless, we retain the property that each of the
underlying primitive actions yields a unique successor state.  Hence,
the solution to the frame problem above can be applied to each of
these states individually, and any properties than can be shown to
hold of every successor state will then be provable consequences of
the nondeterministic action.

The idea can be illustrated by imagining a simple robot that can move
along an unbounded one-dimensional surface. In the deterministic case,
we might have a functional fluent $\position(s)$ and an action
$\exactAdvance(x)$ that changes the position of the robot by some
amount $x$. We could then write the following successor-state axiom
for
$\position$:
\begin{equation}
  \label{eq:ssexactpos}
  \begin{array}{l}
    \Poss(a,s) \supset \position(\dofn(a,s)) = \mbox{}\\
    \qquad \mbox{\bf if\ } \exists{x}.\,a=\exactAdvance(x)\\
    \qquad\qquad \mbox{\bf then\ } \position(s)+x\\
    \qquad\qquad \mbox{\bf else\ } \position(s),
  \end{array}
\end{equation}
assuming that $\exactAdvance$ is the only action affecting
$\position$.%
\footnote{We are taking some liberties here with notation and
the scope
  of variables; the if-then-else formula should be viewed as an
abbreviation for the formula
  \[
  \begin{array}{l}
    \Poss(a,s) \supset \position(\dofn(a,s))=z \\
    \qquad \mbox{} \equiv
    \exists{x}.\,a=\exactAdvance(x) \land z=\position(s)+x \\
    \qquad\qquad \mbox{} \lor
    \lnot\exists{x}.\,a=\exactAdvance(x) \land z=\position(s).
  \end{array}
  \]}

The action $\exactAdvance$ is deterministic---executing it
yields a unique successor state.
Suppose instead we want to model a nondeterministic action
$\noisyAdvance(x)$, which results in the agent moving a distance $y$
which is approximately equal to $x$.  Since we only allow deterministic
primitive actions in our framework, we view $\noisyAdvance(x)$ as the
union of all the possible moves that the agent could actually make,
given that it tries to move $x$.
To capture this, we assume that the language contains primitive actions
of the form $\theAdvance(x,y)$, where $x$ is the {\em nominal\/}
distance the agent is trying to move, by sending an appropriate command
to its on-board effectors, while $y$ is the {\em actual\/} distance
moved.
The agent has control over the nominal distance, but no direct control
over the actual distance
moved. So, in fact, the agent cannot execute an instance of
$\theAdvance(x,y)$ directly.

Nevertheless, we can write a successor-state axiom for $\position$
assuming that it can be changed only by the action $\theAdvance$. The
axiom is the same as above except that the position is changed by the
actual distance moved and the nominal distance is ignored.
\begin{equation}
  \label{eq:sspos}
  \begin{array}{l}
    \Poss(a,s) \supset \position(\dofn(a,s)) = \mbox{} \\
    \qquad \mbox{\bf if\ } \exists{x,y}.\, a=\theAdvance(x,y) \\
    \qquad\qquad \mbox{\bf then\ } \position(s)+y \\
    \qquad\qquad \mbox{\bf else\ } \position(s).
  \end{array}
\end{equation}

Although the agent cannot execute an instance of $\theAdvance$
directly, it can activate its effectors specifying a particular
nominal value for the distance to be moved. This corresponds to
executing a nondeterministic action: execute $\theAdvance(x,y)$ for a
fixed value of $x$ and a nondeterministic choice of $y$.

To specify the execution of such actions we need to expand our
notation for $\dofn$ to allow for more than one potential successor
state.%
\footnote{Remember that $\dofn(\ldots)$ is defined to be a first-order
  term. Hence, it can denote only a single individual, in this case a
  single situation.}
The first step is to allow more complex actions.  Starting with
primitive actions (the ones which already have names in the language),
we form more complex actions by closing off under sequencing and
nondeterministic choice).  In particular, if $\delta_1$ and $\delta_2$
are actions, then so are $\delta_1; \delta_2$ (intuitively, perform
$\delta_1$ and then
$\delta_2$), $\delta_1|\delta_2$ (intuitively,
nondeterministically choose one of $\delta_1$
and $\delta_2$ and perform it),  and $\pi x. \delta_1$ (intuitively,
perform $\delta_1$
for some nondeterministically chosen value of $x$; this is particularly
interesting if $x$ is a free variable in $\delta$).  Given a (complex)
action $\delta$, we follow the approach taken in {\sc golog}
\cite{gologpaper} and
  define $\Do(\delta,s,\succes{s})$ as an abbreviation for a
  situation calculus formula which intuitively reads ``$\succes{s}$ is
  a possible final situation arising from the execution of
  action $\delta$ in situation $s$''. $\Do$ is defined by the
  following recursive expansions:
  \begin{enumerate}
  \item If $\delta$ is a primitive action like $\exactAdvance(x)$ or
    $\theAdvance(x,y)$, then the expansion of $\Do(\delta,s,\succes{s})$
    is:
    \[
    \Do(a,s,\succes{s}) \eqdef
    \Poss(a,s) \land \succes{s} = \dofn(a,s).
    \]
  \item If $\delta$ is of the form $\delta_1\;;\;\delta_2$,
    then the expansion is
    \[
    \Do([\delta_1\;;\;\delta_2],s,\succes{s}) \eqdef
    \exists{s'}.\, \Do(\delta_1,s,s') \wedge \Do(\delta_2,s',\succes{s}).
    \]
  \item If $\delta$ is of the form $\delta_1|\delta_2$,
    then the expansion is
    \[
    \Do(\delta_1|\delta_2,s,\succes{s}) \eqdef
    \Do(\delta_1,s,\succes{s}) \lor \Do(\delta_2,s,\succes{s}).
    \]
  \item If $\delta$ is of the form $\pi{x}.\delta'$,
then the expansion is
    \[
    \Do([\pi{x}.\delta'],s,\succes{s}) \eqdef
    \exists{x}.\,\Do(\delta',s,\succes{s}).
    \]
    This expansion allows for a different successor state for each
    value of $x$.
\end{enumerate}

Now we can formalize a nondeterministic move action
$\noisyAdvance(x)$ that is directly executable by the agent:
\begin{equation}
  \label{eq:noisyadvance}
  \noisyAdvance(x) \eqdef \pi{y}.\,\theAdvance(x,y).
\end{equation}

In this form, $\noisyAdvance$ does not give the agent any
control over its movements since we have not specified any
relationship between the nominal value $x$ asked for and the actual
value $y$ achieved. One reasonable constraint between these two values
might be to assert that there is an absolute bound on the effector's
error. That is, there is a bound on the difference between the value
asked for and the value achieved. This can be captured by asserting
the following precondition axiom for $\theAdvance$:
\begin{equation}
  \label{eq:preadvance}
  \Poss(\theAdvance(x,y),s) \equiv |x-y| \le b,
\end{equation}
where $b$ is the magnitude of the maximum possible error. This axiom
says that it is impossible to execute $\theAdvance$ actions in which
the actual value moved is more that $b$ units different from the value
asked for. When this precondition is combined with the definition of
the nondeterministic action $\noisyAdvance$, it limits the choice of
$\theAdvance$ actions that can arise. Combining this precondition
axiom, the definition of $\noisyAdvance$, and the expansions for $\Do$,
it is not difficult to see that the action $\noisyAdvance(x)$ can
never yield a successor state in which the agent has moved an amount
outside the range $[x-b,x+b]$. More formally we have as a direct
entailment that
\[
\Do(\noisyAdvance(x),s,\succes{s}) \supset |\position(\succes{s}) -
\position(s) - x| \le b.
\]
In Section~\ref{sec:beliefs}, we demonstrate how a probabilistic
relationship can be specified between the nominal and actual values so
as to capture the fact that these two values are correlated---although
the agent's actions do not always have exactly their intended effect
they do have an effect that is correlated with what was intended.

Finally, it should be noted that we have added nothing to the previous
situation calculus story except a few convenient abbreviations. Since
the execution of any complex action ultimately reduces to the execution
of some collection of deterministic actions, we can apply the previous
solution to the frame problem to each of the possible final
situations.

\section{Adding Knowledge}\label{sec:epistemic}
When the agent executes a nondeterministic action like the
$\noisyAdvance$ action described above, the particular deterministic
action that ends up being executed remains unknown to the agent.
Hence, it is clear that in the presence of such actions (and in the
presence of perceptual actions) there needs to be a distinction
between the state of the world and the information the agent has about
the state of the world.  We would like to be able to talk about the
effect of actions like $\theAdvance$ on what an agent or robot knows
or believes about where it is located. For example, doing a
$\noisyAdvance$ decreases the robot's certainty about its position,
but doing a sensing action increases it (in a manner to be described
below). To accomplish this, we need to be able to characterize and
reason about the agent's knowledge and beliefs about the world, i.e.,
the agent's epistemic state, and to distinguish this from the actual
state of the world.

To talk about the actual state of the world in the situation calculus is 
easy: the current situation captures the current state of the world. To 
capture the agent's epistemic state we employ some standard ideas from 
modal logics of knowledge and belief. %
(See \cite{FHMV} for an introduction, and section \ref{sec:fhmv} for a 
discussion of how the approach taken here relates to the model of knowledge 
presented there.) 
In modal logic, an agent has a binary {\em 
possibility relation\/} or
{\em accessibility relation\/} on possible worlds.   The agent is then
said to know a fact $\phi$ in world (or situation) $s$ if $\phi$ is true
at all the worlds the agent considers possible, as captured by the
possibility relation.  It is well known that various properties of
knowledge can be captured by placing constraints on the possibility
relation.  In particular, if we take the possibility relation to be
transitive and Euclidean, the agent has positive and negative
introspection, so it knows what it knows and knows what it does not
know.

These ideas
were first applied in the context of the situation calculus by Moore
\cite{Moore85}, who introduced a binary fluent $\K$ into the language
to capture the possibility relation of modal logic.  That is,
$\K(s',s)$ holds if, in situation $s$, the agent considers situation
$s'$ to be possible.  As in modal logic, we say that the agent knows
$\phi$ in situation $s$ if $\phi(s')$ holds in all situations $s'$
that the agent considers possible in $s$ (i.e., such that $\K(s',s)$
holds).  Moore's approach was extended and integrated with Reiter's
solution to the frame problem by Scherl and Levesque
\cite{scherl-levesque}. The key issue is defining an appropriate
successor-state axiom for $\K$, that specifies how the fluent changes
after doing an action $a$ in some situation $s$. In other words, we
need to specify what has to be true of $\succes{s'}$ for
$\K(\succes{s'},\dofn(a,s))$ to hold.

Scherl and Levesque do this by dividing actions into ordinary ones (like 
moving) and knowledge-producing ones intended to change only the $\K$ 
fluent. However, their characterization entails knowledge of the action 
executed; i.e., the agent always comes to know what action (of either type) 
was executed. We need to model the case where nondeterministic actions are 
being executed by the agent, and the agent does not come to know the exact 
action being executed. For example, if the action happens to be
$\theAdvance(x,y)$, we do not expect the agent to know that this
action was performed, since that would mean knowing the actual
distance $y$ moved. Rather, we expect the agent to know only the
nominal value $x$ that it tried to move and whatever constraints this
places on $y$.   That is, as a result of performing this action,
the agent learns just that it tried to move $x$.  

Another way of thinking about this is that,
after doing $\theAdvance(x,y)$,
the agent knows only that $\theAdvance(x,y')$ was performed for
some compatible value $y'$.
To formalize this idea, we use a special predicate $\ObsIn(a,a',s)$,
meaning $a$ and $a'$ are observationally
indistinguishable, i.e.,
cannot be distinguished by the agent. We assume that as part of the
background theory, the user specifies a collection of {\em
  observation-indistinguishability axioms\/} characterizing this
predicate.
\begin{defn}[Observation-Indistinguishability Axioms]
  \label{def:oeaxioms}
  {\em Observation-indistinguishability axioms\/} are axioms of the
form
  \[
  \ObsIn(a,a',s) \equiv \phi(a,a',s),
  \]
  one for each action $a$. Here $\phi$ is a formula that characterizes
  the relationship between actions $a$ and $a'$ that makes them
  indistinguishable in situation $s$. These axioms specify the set of actions
  $a'$ that the agent has no ability to discriminate from action $a$
  in situation $s$. That is, in situation $s$ the agent cannot tell if
  action $a$ or $a'$ was executed.
\end{defn}
For ordinary actions, i.e., primitive actions for which the agent
knows when they are performed and knows all of their consequences, this
axiom would say that $\ObsIn$ holds iff $a=a'$. For example, if
$\pickup(x)$ is an ordinary action, we would have
\[
\ObsIn(\pickup(x),a',s) \equiv a'=\pickup(x).
\]
For actions like $\theAdvance(x,y)$, however, the axiom would be
\begin{equation}
  \label{eq:obsadvance}
  \ObsIn(\theAdvance(x,y),a',s) \equiv \exists{y'}.\,
  a'=\theAdvance(x,y').
\end{equation}
This says that, for example, the agent cannot distinguish between the
executions of the actions $\theAdvance(3.0,3.2)$ and
$\theAdvance(3.0,2.8)$, since both involve attempting to move the same 
3.0 units.

Using $\ObsIn$, we can specify a new successor-state axiom for $\K$
that correctly handles actions like $\theAdvance(x,y)$.
\begin{defn}[$\K$'s Successor-State Axiom]
  \label{def:ssK}
  \[
  \begin{array}{l}
    \Poss(a,s) \supset \K(\succes{s'},\dofn(a,s)) \\
    \qquad \mbox{} \equiv
    \exists{a',s'}.\; \succes{s'} = \dofn(a',s') \land \Poss(a',s') \land
    \ObsIn(a,a',s)  \land \K(s',s)
  \end{array}
  \]
\end{defn}
This axiom says that when the agent is in situation $\dofn(a,s)$
(after having executed action $a$ in situation $s$), the set of
$\K$-related situations (that define its state of knowledge) are
precisely those situations that are the result of executing an
observationally indistinguishable action $a'$ from some situation $s'$ that
satisfies the preconditions of $a'$ and was $\K$-related to the
previous situation $s$.
Intuitively, when the agent is in $s$, as far as it knows it could be in 
any situation $s'$ that is $\K$-related to $s$. Hence, as far as it knows 
it could have executed $a$ in any of these situations, and all of the $a$ 
successors of these situations must be in its new knowledge set. 
Moreover, it does not even know if it in fact executed $a$; it could 
have executed any $a'$ indistinguishable from $a$, so the $a'$ successors 
of these situations must also be in its new knowledge set. However, it does 
know that an action indistinguishable from $a$ was successfully executed. 
Hence, it can eliminate from its knowledge set the successors of those 
situations that fail to satisfy the preconditions of some $a'$.
It is not hard to show 
that for ordinary actions like $\pickup(x)$, which are
observationally indistinguishable only from themselves, this axiom
reduces
to the axiom given by Scherl and Levesque 
\cite{scherl-levesque}.\footnote{See Section~\ref{sec:kpa} for how this axiom
also correctly handles Scherl and Levesque's knowledge producing actions.}

As we said earlier, we define knowledge to be truth in all worlds the
agent considers possible.  For example, to say that the agent knows in 
situation $s$ that 
object $x$ is not broken, we would state
$\forall s'.\,K(s',s) \supset \lnot\broken(x,s').$
It is convenient to introduce special syntactic machinery to express 
this condition:
\begin{defn}[$\Know$]
  \label{def:know}
Let $\snow$ be a special situation term.
  We write $\Know(\phi[\snow],s)$ to indicate that the agent knows
  $\phi[\snow]$ in situation $s$. 
Thus, $\Know(\phi[\snow],s)$ is an abbreviation for the
formula $ \forall{}s'.\,\K(s',s) \supset \phi[\snow/s']$, 
where $s'$ is some variable not appearing anywhere in $\phi$, and 
$\phi[\snow/s']$ is the result of replacing all free\footnote{An 
occurrence of $\snow$ is considered to be bound if it appears within the scope 
of $\Know$ and free otherwise.} occurrences of $\snow$ in $\phi$ by $s'$. 
\end{defn} For example, $\Know(\lnot\broken(x,\snow),s)$ is an abbreviation 
for the above formula. Notationally, it is often convenient to suppress the 
$\snow$ term and simply write $\Know(\lnot\broken(x),s)$, assuming that it 
will be clear from context where the implicit situation variable $s'$ needs to 
be inserted.

\begin{example}
  Given the axioms above, it is easy to show that after doing a
  $\noisyAdvance(x)$, the agent will know that its current position is
  its previous position plus $x$, give or take $b$ units. More
  precisely, let us assume that we have a special function $\prev(s)$
  which for situations other than $S_0$ denotes the situation
  immediately before $s$.%
  \footnote{We can characterize this function by the axiom
    $\prev(\dofn(a,s)) = s$.} %
  Then these formulas entail
  \[
  \begin{array}{l}
    \Do(\noisyAdvance(x),s,\succes{s}) \\
    \qquad\mbox{} \supset
    \Know(\position(\snow) \in
    [\position(\prev(\snow)) + x - b, \position(\prev(\snow)) + x
+b],\succes{s})
  \end{array}
  \]
  or, in more detail,
  \[
  \begin{array}{l}
    \Do(\noisyAdvance(x),s,\succes{s}) \\
    \qquad \mbox{} \supset
    \forall{\succes{s'}}.\,\K(\succes{s'},\succes{s}) \supset
    \exists{a',s'}.\; \succes{s'} =\dofn(a',s') \\
    \qquad\qquad \mbox{} \land
    (\position(s')+x - b) \leq \position(\succes{s'}) \leq
    (\position(s')+x +b).
  \end{array}
  \]

  Briefly, any situation $\succes{s}$ that is the result of executing
  $\noisyAdvance(x)$ must have been the result of executing 
  $\theAdvance(x,y)$ for some $y$. According to
(\ref{eq:preadvance}),
$\theAdvance(x,y)$ is executable only if $|x-y|\leq b$.
Hence, from (\ref{eq:sspos}) we know
  that in
$\succes{s}$, the agent's position is within $b$ of its position at $s$
plus $x$.  By Defn.~\ref{def:ssK} and (\ref{eq:obsadvance}), this holds 
in all of the situations that are $\K$-related to $\succes{s}$ as well:
all of these situations are also the
result of executing $\theAdvance(x,y)$ for some $y$ with $|x-y|\leq b$.
Hence, the agent knows this in $\succes{s}$.

It is not hard to show that the agent would also know that this property holds 
before executing the $\noisyAdvance(x)$ action, that is, that the formula
$\Know(\psi,s)$ is entailed, where $\psi$ is
  \[
  \begin{array}{l}
    \forall{\succes{s}}.\Do(\noisyAdvance(x),\snow,\succes{s}) \\
    \qquad\mbox{} \supset
    \position(\succes{s}) \in
    [\position(\snow) + x - b, \position(\snow) + x +b].
  \end{array}
  \]

  Observe that a noisy advance decreases positional certainty, in
  that even if the agent knows its current position precisely, after
  doing a noisy advance it will know its position only to within $b$
  units.  Similarly, if the agent knows its current position to within
  $a$ units it will know its position only to within $b+a$ units after
  doing a noisy advance. In particular, multiple advances add to the
  agent's uncertainty, so the following formula is also entailed.
  \[
  \begin{array}{l}
    \Do(\noisyAdvance(x)\;;\;\noisyAdvance(y),s,\succes{s})\\
     \qquad\mbox{} \supset 
   \Know(\position(\succes{s}) \in [\position(\prev(\prev(\snow))) + x + y - 2b,\\
    \qquad\qquad\mbox{} \position(\prev(\prev(\snow))) + x + y + 2b],\succes{s})
  \end{array}
  \]
\end{example}

\subsection{Sensing Actions}
Sensing actions can also be handled within the account we have
developed so far. Sensing actions are actions that are executed
primarily for the change they produce on the $\K$ fluent; pure sensing
actions would affect no other fluent, and would appear only in the
successor-state axiom of the $\K$ fluent.

For example, imagine that the robot has a sensor that senses its
current location in our 1-dimensional world. The sensor is subject to
error, so the value it returns is unlikely to be the exact location.
We can model the act of using such a sensor with an action
$\thePos(x,y)$ analogous to $\theAdvance(x,y)$. In this case, the
nominal value $x$ would be the value appearing on the sensor when
the sensing action is executed,
and the actual value $y$ would be the
true position. As with $\theAdvance$, we assume that the agent cannot
observe the actual value. This gives the following
observation-indistinguishability axiom for $\thePos$:
\begin{equation}
\label{eq:obspos}
\ObsIn(\thePos(x,y), a',s) \equiv \exists y'.\, a'=\thePos(x,y').
\end{equation}

What makes $\thePos(x,y)$ carry information is the fact that the
true position $y$
places constraints on the value of $x$ that can be read from the sensor.
This is captured by the precondition axiom for this action,
which is
\begin{equation}
  \label{eq:prepos}
  \Poss(\thePos(x,y),s) \equiv y= \position(s) \land |x-y| \le c.
\end{equation}
In this case, we have simply asserted that there is a bound on the
error in the value read from the sensor: it must be within $c$ units
of the true value.

The main difference between $\thePos$ and $\theAdvance$ is that with
$\thePos$ the robot does not have the ability to select either the
nominal value read $x$ or the actual value $y$.  The first of these
is chosen nondeterministically, and depends on the error in that
activation of the sensor; the second (by virtue of the above
precondition axiom)
is determined
by the actual position of the
robot. Thus, we assume that the robot merely gets to execute the
nondeterministic action $\noisyPos$ which takes no arguments:
\begin{equation}
  \label{eq:noisypos}
  \noisyPos \eqdef \pi{x,y}.\,\thePos(x,y).
\end{equation}

\begin{example}
  Given the axioms above, we can show that after doing a $\noisyPos$,
  the agent will know what its current position is to within $c$ units.
  More precisely,
\[
\begin{array}{l}
  \Do(\noisyPos,s,\succes{s}) \\
  \qquad \mbox{} \supset
  \exists{x}.\,\Know(\position \in [x-c,x+c],\succes{s}),
\end{array}
\]
  or, in more detail,
  \[
  \begin{array}{l}
    \exists{x}\exists{y}\,\succes{s}=\dofn(\thePos(x,y),s) \\
    \qquad \mbox{} \supset
    \exists{x}\forall{\succes{s'}}.\,\K(\succes{s'},\succes{s}) \supset
    (x-c) \leq \position(\succes{s'}) \leq
    (x+c).
  \end{array}
  \]
is entailed by these axioms.

Once again, any situation $\succes{s}$ that can arise from executing
$\noisyPos$ must be the result of executing a particular instance of
$\thePos(x,y)$.
According to (\ref{eq:prepos}),
these instances are executable only if
$y=\position(s)$ and
$x\in[\position(s)-c,\position(s)+c]$. Hence, there is a fixed
value $x$ (the value that will be read from the sensor) such that
$\position(s) \in [x-c,x+c]$. Furthermore, by (\ref{eq:sspos}) we
see that $\thePos$ actions do not affect the current position, so
$\position(\succes{s})=\position(s)$ and the bound holds in
$\succes{s}$ as well.

Now if we examine all situations $\succes{s'}$ $\K$-related to
$\succes{s}$ we see that (1) they must be the result of executing an
observationally indistinguishable action $\thePos(x',y')$ in a situation $s'$
that is $\K$-related to $s$, which by (\ref{eq:obspos}) means
$x'=x$, and (2) $\thePos(x',y')$ must be possible in $s'$, which by
(\ref{eq:prepos}) means $y'=\position(s')$ and $y'\in[x'-c,x'+c]$.
Putting these together we see that $\position(s') \in [x-c,x+c]$.
Furthermore, again by (\ref{eq:sspos}), we obtain
$\position(\succes{s'}) \in [x-c,x+c]$. Hence, the position in every
$K$-related state lies in a fixed range, so the agent must know this.
\end{example}

There are a couple of points worth noting. First, if the agent has
very incomplete knowledge of its position, then its knowledge of its
position will increase after reading its sensor. This increase of
knowledge arises from the pruning away of $K$-related situations that
have divergent values of $\position$. In particular, in situations
with extreme values of $\position$ there will be no observationally
indistinguishable action $\thePos(x,y')$ that can be executed (the fact that
$y'$ must equal the situation's $\position$ value will violate the
bound between $y'$ and the fixed value of $x$).

Second, if the agent has strong knowledge of its position, then that
knowledge will not degrade after reading its sensor. For example,
suppose that
the agent knows the exact value of its current position. This means
that $\position$ has the same value in all situations $K$-related to
the current situation. In this case, the agent will continue to know
its exact position after reading its sensor. This arises from the fact
that all situations that are $K$-related to the successor state must
arise from the execution of a $\thePos$ action in a situation that was
$K$-related to the current state. Since, $\thePos$ actions do not
change the value of $\position$, they will all continue to share the
same value of $\position$.

\section{Likelihood and Degree of Belief}
\label{sec:beliefs}
Asserting bounds on the difference between the nominal values and the
actual values in actions like $\thePos(x,y)$ represents a rather weak
model of action uncertainty. Suppose, for example, we have a sensor
with an error bound of $c = 2$, and we make a number of readings of a
particular fluent using the sensor, all of which are clustered around
the value 3. For concreteness, suppose they are all between 2.8 and
3.1. As far as {\em knowledge\/} goes, all the agent will be able to
conclude is that it knows the fluent to have a value in the range
[1.1,4.8]. Getting numerous readings of 3 will not change this
knowledge. Yet, even if the agent is using a cheap sensor, we might
hope that getting many such readings would increase the agent's {\em
  degree of belief\/} that the true value of the fluent is indeed close to 3.

To formalize these intuitions, we first quantify the notion of
possibility (as captured by the $\K$ fluent) by associating with each
world the agent considers possible the agent's degree of belief that
that is the actual world.  We can think of this degree of belief as a
subjective probability.  We then consider how these degrees of belief
change over time, as a result of actions being performed.

\paragraph{Degrees of Belief.}
It is convenient to capture degree of belief by first associating with
each situation a {\em weight}, and then taking the degree of belief to
be the normalized weight (since, like probability, we want the degree of
belief in the whole space to be 1).
We capture the weight by using a new functional fluent
$\pr(s',s)$, analogous to
$\K(s',s)$. This function denotes the relative weight given to
situation $s'$ by the agent when it is in situation $s$.

We expect weights to be non-negative and that
all situations considered impossible will be given weight 0.
The following constraints, which we assume is part of the background
action theory ensures that this is indeed the case initially:
\begin{equation}
  \label{eq:prob}
  \begin{array}{l}
    \forall{s}.\, \pr(s,S_0) \geq 0\, \land\, 
     \lnot \K(s,S_0) \supset \pr(s,S_0) = 0.
  \end{array}
\end{equation}
As we shall see, the successor-state axiom for $p$ ensures that these
constraints hold at all times.

We take the agent's degree of belief in $\phi$ to be the total weight of
all the worlds he considers possible where $\phi$ holds, normalized by
the total weight of all worlds he considers possible.
Thus, we are restricting to {\em discrete\/} probability distributions
here, where the probability of a set can be computed as the sum of the
probabilities of the elements of the set.%
\footnote{
  This is in contrast to continuous probability distributions, 
  in which the probability of each point is 0, so the probability of a
  set cannot be computed as the sum of the probabilities of the
  elements in the set. Instead we must integrate over the set.} We
introduce special notation for this, just as we did for $\Know$.
\begin{defn}[$\bel$]
  \label{def:bel}
Let $\snow$ be a special situation term.
  We write $\bel(\phi[\snow],s)$ to denote the agent's degree of belief
in $\phi[\snow]$ when it is in situation $s$.
This is an abbreviation for the term
  \[
  \left. {\sum}_{\{s': \phi[\snow/s']\}} \pr(s',s) \right/ {\sum}_{s'}
\pr(s',s).
  \]
As with $\Know$, we sometimes suppress the $\snow$ terms. In 
Appendix~\ref{app:belief}, we show how the summations in this formula can 
be expressed using second-order quantification. A
  logical consequence of this formalization is that $\bel(\cdot,s)$ is
  a probability distribution over the situations $\K$-related to $s$.
\end{defn}

\paragraph{Action-Likelihood Functions.}
Next we have to define a successor-state axiom for the $p$ fluent,
that is, we need to consider how the agent's probability distribution
changes over time.  It will not be necessary to characterize how belief 
changes, since belief is defined in terms of $p$.

Suppose the agent performs an action $a$ in some
situation $s$.  If $a$ is an ordinary deterministic action like
$\pickup(x)$, then the agent knows it will end up in a unique situation
$\dofn(\pickup(x),s)$.  Intuitively, in this case, the
probability that it ascribed to $s$ will be transferred to 
$\dofn(\pickup(x),s)$. Similarly, this is true for all the situations that the 
agent considers possible at $s$.

But suppose that $a$ is an action like $\thePos(x,y)$: sensing a value of $x$ 
when the true position is $y$. In this case, the probability the agent 
ascribes to $\dofn(a,s)$ will not simply be what is transferred from $s$; how 
likely the action $a$ is must also be taken into account. As discussed above, 
$\thePos(x,y)$ is not an action that an agent can execute directly: it executes
something like $\noisyPos$, and one of the $\thePos$ actions is selected 
nondeterministically. But some of these are more likely to be selected than 
others. For example, if the true position in situation $s$ is $3.0$, we expect 
that $\thePos(3.0,3.0)$ is far more likely to be executed than
$\thePos(30.0,3.0)$.  Consequently,  
the probability the agent ascribes to the situation 
$\dofn(\thePos(3.0,3.0),s)$ should be greater than what it ascribes to 
$\dofn(\thePos(30.0,3.0),s).$

To specify these probabilities, we use what we call an {\em 
action-likelihood\/} function. We add a special function $\lhood(a,s)$ to our 
language, used to denote the probability (assigned by the agent in situation 
$s$) of primitive action $a$ being selected among all its possible 
observationally indistinguishable alternatives. Thus, for example,
$\lhood(\thePos(3.1,3.0),s)$ is the likelihood that the agent would sense 3.1 
when its true position was 3.0 in situation $s$.
Note that if $a$ is an ordinary deterministic action like $\pickup(x)$, then 
its only observationally indistinguishable alternative is $a$ itself, so,
as long as $a$ is possible in $s$, we would normally have $\lhood(a,s) = 1$.

The error profile of the various effectors and sensors available to
the agent is clearly application dependent, and it is this error
profile that is captured by the $\lhood(a,s)$ function. Hence, we
assume that the user specifies a collection of axioms characterizing
$\lhood$ for each action $a$.
\begin{defn}[Action-Likelihood Axioms]
  \label{def:lkaxioms}
  {\em Action-likelihood axioms\/}  are axioms of the form
  \[
  \lhood(a,s)=z  \equiv \phi(a,s).
  \]
  Here $\phi(a,s)$ is a formula that characterizes the
  conditions under which
action $a$ has likelihood $z$ in situation $s$. We will see some example 
axioms below. 
\end{defn}

With this machinery, we can give the successor-state axiom for $\pr$,
and hence characterize how belief changes as the result of performing
an action. Suppose situation $\succes{s'}$ is of the form
$\dofn(a',s')$, that is, it is the result of doing action $a'$ in
situation $s'$. Then the weight that the agent assigns to
$\succes{s'}$ in $\dofn(a,s)$ is the product of the weight it assigns
to being in situation $s'$ and the likelihood of $a'$ having been
performed: the former is 0 unless $K(s',s)$ holds, in which case it is
$\pr(s',s)$; the latter is 0 unless $a'$ is observationally
indistinguishable {}from $a$ and possible in situation $s'$, in which
case it is $\lhood(a',s')$.
\begin{defn}[Successor-State Axiom for $\pr$]
  \label{def:sspr}
  The successor-state axiom for $\pr$ is
  \[
  \begin{array}{l}
    \Poss(a,s) \supset \pr(\succes{s'},\dofn(a,s)) = \\
    \qquad \mbox{\bf if\ }
    \exists{a',s'}.\; \succes{s'} = \dofn(a',s') \land \Poss(a',s')
    \land \ObsIn(a,a',s)  \land K(s',s)\\
    \qquad\qquad \mbox{\bf then\ }
    \pr(s',s) \times \lhood(a',s') \\
    \qquad\qquad\mbox{\bf else\ } 0.
  \end{array}
  \]
\end{defn} Note that the axiom captures the fact that any situation that is 
not $K$-related to the successor state will be assigned probability 0: the 
precondition to the {\bf then} clause is precisely the condition required for 
a situation to be $\K$ related to the successor state (Defn.~\ref{def:ssK}). 
It also ensures that when $\Poss(a',s')$ is false, the weight assigned to the 
situation $do(a',s')$ will be 0.\footnote{This can also be achieved by 
eliminating $\Poss$ from this axiom and requiring $\lhood(a',s')$ to be 0 when 
$\Poss(a',s')$ is false.}

As a concrete example of how action-likelihood axioms can be used, consider 
an effector action like $\theAdvance(x,y)$, where the 
agent chooses the nominal value $x$ and the actual value $y$ is 
nondeterministic.
Suppose the correlation between $x$ and $y$ is described by a {\em linear
  Gaussian} model: the actual value $y$ is the nominal value
$x$ plus some random noise that has a normal distribution.
This means that $y-x$, the difference between the
actual movement and the nominal movement, is normally distributed, with
some mean $\mu$ and variance $\sigma^2$.
If we further assume that the effector is calibrated so that there is no
systematic bias, then $\mu= 0$, and the distribution of $y-x$ is
characterized by $\norm((y-x)/\sigma)$, where
$\norm(z)$ is the standardized normal distribution
with mean 0 and variance 1.%
\footnote{There is a slight subtlety here.  The normal distribution
  is, of course, a continuous distribution, so the probability of a
  single real number is always zero.
  Only non-trivial intervals of the real line have positive
  probability.
  We are assuming that all probability functions are discrete here, so
  we really need to consider an approximation to the normal
  distribution. In examples like this, we can do so by treating action
  parameters as finite precision numbers. In this case, we interpret
  $\norm(z)$ as being the integral of the normal density function over
  the range of precision of $z$.  For example, if the actuator allows
  the agent to ask for only movements like 3.1 with 2 digits of
  precision, the range denoted by 3.1 could be 3.05--3.15, and
  $\norm(3.1)$ would be the value of the integral of the normal
  density function over this range. Given some fixed precision, a
  table of values for $\norm$ can be computed and added to the domain
  theory as a set of equations.
  {}From here on in, when we write $\norm$, we actually mean the
  discrete probability distribution that approximates $\norm$.}

Given these assumptions,
the following axiom describes the action:
\begin{equation}
  \label{eq:lkadvance}
\lhood(\theAdvance(x,y),s) = \norm((y-x)/\sigma).
\end{equation}

 This axiom captures the linear Gaussian error model,
asserting
that the probability of the effector moving $y$ units when asked to move
$x$ units falls off like the normal distribution as the difference
between these two values increases.
We can do something similar for sensor actions.
If $x$ is the value read and $y$ is the actual value,
we might once again assume that their difference $x-y$ is normally
distributed, say with variance ${\sigma'}^2$ and mean 0.  (Again,
assuming a mean of 0 implies that there is no systematic bias in the
reading.)
This gives us the following action-likelihood axiom:
\begin{equation}
  \label{eq:lkpos}
  \lhood(\thePos(x,y),s) =
\left\{
\begin{array}{ll}
\norm((y-x)/\sigma') &\mbox{if $\position(s) = y$}\\
0                    &\mbox{if $\position(s) \ne y$}
\end{array}
\right.
\end{equation}
which specifies that the sensor's error profile has a normal
distribution, but with perhaps a different variance ${\sigma'}^2$.
In either case, we could have also insisted that $\lhood(a,s)$ be 0 when 
$\Poss(a,s)$ is false (for example, when the difference between the nominal and 
actual value exceeds some bound), but as noted above, this is not necessary
because of the way we have written the successor-state axiom for $\pr$. 

\begin{example}
  \label{ex:contextLikelihood}
  When specifying the likelihood function we can encode complex
  context-dependent error profiles. Suppose have a fluent $\slippery$
  that is true of situations where the robot is located on a slippery
  surface. We could account for a context-dependent effect on the
  robot's motion effectors with an axiom like
  \[
  \begin{array}{l}
    \lhood(\theAdvance(x,y),s) = \mbox{} \\
    \qquad \qquad \mbox{\bf if\ } \slippery(s) \\
    \qquad \qquad\qquad \mbox{\bf then\ } \norm((x-y)/2.4)\\
    \qquad \qquad\qquad \mbox{\bf else\ } \norm((x-y)/0.5)\\
  \end{array}
  \]
  which indicates that the probability of error increases
  significantly when the surface is slippery (the standard deviation
  increases). To the best of our knowledge, no other formalism 
for representing the uncertainty associated with sensors and effectors
allows this sort of context-dependent error profile to be expressed.

Similarly, suppose that the effector's error was cumulative. That is,
the further the robot tried to move the greater the error. Then an
action-likelihood axiom like the following could be used:
  \[
  \lhood(\theAdvance(x,y),s) =
\norm((x-y)/(cx))    
  \]
  This axiom specifies that the standard deviation is some factor,
  $c$, of the nominal distance the agent is attempting to move.
\end{example}

\section{Summary of the Formalization} \label{sec:summary} 

This concludes the formal details of our approach. But before looking at some 
its properties, let us summarize the components. To incorporate noisy sensors 
and effectors as well as degrees of belief into the situation calculus, we 
start with the situation calculus language from before, but add two additional 
fluents, $\K$ and $\pr$, a new distinguished predicate $\ObsIn$ and function 
$\lhood$. We need to extend the basic theory of action (see 
Section~\ref{sec:framesol}) to what we will now call an {\em extended theory 
  of action}, consisting of the following axioms:
\begin{itemize}
  
\item axioms describing the initial situation, as before, but now
  including Axiom~\ref{eq:prob} for $K$ and $\pr$ in $S_0$;
  
\item action-precondition axioms, as before;
  
\item successor-state axioms, as before, but now including one for $K$
  (Defn.~\ref{def:ssK}) and one for $\pr$ (Defn.~\ref{def:sspr});
  
\item unique-name axioms for the primitive actions, as before;
  
\item observation-indistinguishability axioms
  (Defn.~\ref{def:oeaxioms}) characterizing $\ObsIn$, one for each
  action;
  
\item action-likelihood axioms (Defn.~\ref{def:lkaxioms})
  characterizing $\lhood$, one for each action;

\item various abbreviations for convenience, including $\Know$, $\bel$,
  and $\Do$;

\item axioms (given in Appendix~\ref{app:belief}) to ensure that
  $\bel$ is always a probability distribution.
  
\item the foundational domain-independent axioms. These need to be
  modified to take into account that because of its epistemic
  alternatives, $S_0$ is no longer the only initial situation
  \cite{levesque-lakemeyer:97}. However, nothing we do here hinges on
  these details. 

\end{itemize}
As we shall show, all of the expected properties are then logical
consequences of these axioms. In taking this approach, we obtain the
full expressiveness of the situation calculus for reasoning about
action, a simple solution to the frame problem, as well as an
expressive language for dealing with the uncertainty arising from
sensors and effectors.

\section{The Formalism at Work}\label{sec:examples} 

We have presented a particular model for noisy actions and their
effects on an agent's beliefs. Our model involves using
nondeterministic actions with probabilities attached to the various
nondeterministic choices. This gives a model of actions that is
essentially identical to standard probabilistic actions, where an
action yields various successor states with varying probabilities.
Although this is a fairly standard model, it is difficult to defend
its correctness in any formal way. The best that can be done is to
demonstrate that the model behaves in a way that matches our
intuitions and properly replicates other more specialized approaches.

\subsection{Knowledge-producing actions}\label{sec:kpa}

In the Scherl and Levesque approach, in addition to ordinary actions like 
$\pickup(x)$, there are assumed to be special knowledge-producing action
associated with various fluents. For example, to open a safe by dialing its 
combination, an agent needs to know what the combination is. If the 
combination of the safe is written on a piece of paper, there might be a 
knowledge-producing action $\readcombi$ whose effect is to change the the 
state of knowledge to make the value of the $\combi$ fluent known.

We can model this type of action as follows.  Assume we have a
primitive action $\readPaper(z)$ which is the action of reading the
number $z$ on the piece of paper.  We assume that this action only
happens if $z$ is the combination of the safe, which we model using a
precondition axiom.
\[
\Poss(\readPaper(z),s) \equiv z=\combi(s).
\]
In this case, there is no distinction between the number observed on
the paper and the actual value, and so this action is observationally
indistinguishable only from itself:
\[
\ObsIn(\readPaper(z),a,s) \equiv a=\readPaper(z).
\]
Finally, the agent cannot choose which number to see,
it can only choose whether or not to see what is written on the paper.
Consequently, we define $\readcombi$ by
\[
  \readcombi \eqdef \pi{z}.\,\readPaper(z).
\]
It is then easy to show, that after reading what is on the paper, the
agent knows the combination to the safe.  That is, the following is
entailed:
\[
\Do(\readcombi,s,\succes{s}) \supset \exists{z}.\Know(z=\combi,\succes{s}).
\]
This approach clearly generalizes to handling knowledge-producing actions 
which tell the agent the truth value of some fluent. One nice advantage of the 
approach here is that, unlike in Scherl and Levesque, it is not necessary to 
mention by name the knowledge-producing actions in the successor-state axiom 
for $K$, which means that it is not necessary to know them all in advance.

\subsection{The FHMV Model of Knowledge}\label{sec:fhmv}

To further understand our approach to modeling the agent's knowledge
and belief in the situation calculus, it is useful to compare it with
the approach to modeling multi-agent systems taken 
in \cite{HF89,FHMV} (and expanded in \cite{HT} to allow probabilistic
actions). In this approach, agents are always assumed to be in some
{\em local state}.  Besides the agents, there is an {\em environment},
which at any given time is also in a particular state. The
environment's state captures everything that is relevant to the
description of the system and not captured by the agent's state. Thus,
if there is only one agent (as is the case here), the system's {\em
  global state\/} at any time can be described by a pair $(s_e,s_a)$,
where $s_e$ is the environment's state and $s_a$ is the agent's state.
The system's global state changes as a result of actions being
performed. These actions may change either (or both) of the components
of the global state. For example, a sensing action that did not change
the environment could be modeled as changing only the agent component
$s_a$ of the global state.

What we are doing here using the situation calculus can be modeled
within this framework by mapping situations to global states. That is,
we map every situation $s$ to a pair $(s_e,s_a)$. This mapping can be
constructed in a variety of ways. Perhaps the easiest is to take
$s_e$, the environment's state, to consist of the values of the
fluents (other than $\dofn$, $K$, and $\pr$) in $S_0$, together with
the sequence of primitive actions used to generate the situation.
Similarly, the agent's state, $s_a$, can consist of an initial state
that describes the agent's beliefs in $S_0$, together with a sequence
that records what the agent learns as a result of performing the
actions described in the corresponding environment state.  In our
approach, the agent learns only that an observationally
indistinguishable action was executed; so an appropriate sequence for
the agent state can be generated by replacing each action $a$ in the
corresponding environment sequence by a unique canonical
representative from the set of actions that are observationally
indistinguishable to $a$ (with respect to the situation that $a$ was
executed in). We remark that Lakemeyer independently used this
approach to give semantics to knowledge and only knowing in the
situation calculus \cite{Lakemeyer96} (also see
\cite{levesque-lakemeyer:97}). This method of mapping situations to
global states captures the implicit situation calculus assumption that
the agent remembers all of the actions that have been performed
(modulo observationally indistinguishability). It is also obvious that
from any global state $(s_e,s_a)$ we have sufficient information to
reconstruct the situation $s$ that gave rise to it.

With this view of situations, the definition of $\dofn$ for the FHMV
model is immediate. If $(s_e,s_a)$ is the global state corresponding to
situation $s$, then $\dofn(a,(s_e,s_a))$ is the new global state
$(s'_e,s'_a)$ where $s'_e$ is the result of appending $a$ to $s_e$,
and $s'_a$ is the result of appending a canonical member of $\{x:
\ObsIn(a,x,s)\}$ to $s_a$. Similarly, there is an obvious definition
for $K$ as well, which is the same as that taken in \cite{HF89,RK}:
$K((s'_e,s'_a),(s_e,s_a))$ holds precisely if $s'_a=s_a$, i.e., the
agent has the same local state. Notice that this makes $K$ an
equivalence relation, that is, reflexive, symmetric, and transitive.
Finally, we can model $\pr$ by defining $\pr((s'_e,s'_a),(s_e,s_a))$
to be equal to $\pr(s',s)$ where $s'$ is the situation corresponding
to the global state $(s'_e,s'_a)$. By normalizing summations over
$\pr$ we obtain a distribution that defines the agent's probabilistic
beliefs when it is in global state $(s_e,s_a)$.

Besides helping to clarify our approach to modeling the agent's
epistemic state, the discussion above also shows that the situation
calculus is not fundamental. We could have taken essentially the same
approach using a modal logic (the FHMV model is formalized in terms of
modal logic) instead.

\subsection{Bayesian conditioning}
A standard model for belief update in the light of noisy information
is Bayesian conditioning.  The standard Bayesian model assumes two
pieces of probabilistic information: a prior distribution $\Pr(t)$ on
the value $t$ being sensed, and a conditional distribution $\Pr(x|t)$
that gives the probability of sensing $x$ given that the true value is
$t$.  Furthermore, the standard model requires the assumption that the
value read {}from the sensor is dependent only on the true value, and
is thus independent of other factors given this value.

Bayes' Rule is now applied to obtain a posterior probability
$\Pr(t|x)$ over the values of $t$, given that the sensor read the value
$x$: $\Pr(t|x) = \Pr(x|t)\Pr(t)/\Pr(x)$. The denominator---the prior
probability of reading the value $x$---is the only unknown expression,
but it can be easily computed. Since $\sum_{t'} \Pr(t'|x) = 1$, we
must take $\Pr(x)$ to be the normalizing factor $\sum_{t'}
\Pr(x|t')\Pr(t')$. Since this normalizing factor is independent of $t$,
we see that the key factor in determining the posterior probabilities
is the numerator $\Pr(x|t)\Pr(t)$, which describes the relative
probability of different values of $t$ given the observation $x$.

Two significant assumptions are typically made when applying Bayesian
conditioning. The first is that the world does not change while the
observation is being made, and the second is that conditioning on a
value (like $x$ above) is the same as conditioning on the event of
observing $x$. 

In general, there need not be a distinction between sensing actions,
that only affect the agent's beliefs, and ordinary actions that affect
the agent's environment: in our formalism it is quite possible to
define actions that change the world in the course of making an
observation. For such actions, conditioning the agent's beliefs only
on the value sensed (as is done above) does not adequately describe
the agent's new belief state. It is necessary to also take into
account the effect of the other changes caused by the action (this is
the so-called ``total evidence'' requirement of Bayesian
conditioning). In our formalism the agent's new beliefs (as described
by the updated $\pr$) is in fact affected by both the sensed value and
any other changes caused by an action.

To understand the second assumption, imagine that there are two
fluents, $\light$ and $\book$, where $\light(s)$ holds if the light is
on in situation $s$ and $\book(s)$ holds if there is a book in the
room. Initially, say that the agent considers the four possible
situations characterized by these two fluents as equally likely.
Suppose the agent does a sensing action that detects whether or not
the book is in the room, but only if the light is on; when the light
is off nothing can be determined. If the sensing action does in fact
detect the book, then the agent should ascribe probability 1 to the
light being on. However, this is not the probability obtained by
simply conditioning the initial uniform distribution on $\book$. The
probability that the light is on given that the book is in the room is
$1/2$, but the probability that the light is on given that the book
was observed to be in the room is 1.  The naive application of Bayes'
rule implicitly assumes that conditioning on the book being in the
room is equivalent to conditioning on observing the book. 

In such
examples, we are dealing with a sensing action that is dependent on
more than just the value (or truth) of the fluent being sensed. It
depends on other features of the situation as well. As demonstrated in
Example~\ref{ex:contextLikelihood}, our formalism can model
context-dependent sensing actions, and the agent's new beliefs will
reflect both the values sensed as well as what can be learned about the
context from that value. In this case, sensing the value true for the
fluent $\book$ also allows us to learn that $\light$ is true.

Our formalism does not force us to make these two assumptions, but
when we do our approach updates beliefs in a manner identical to
standard Bayesian conditioning. We demonstrate this in the next
section.

\subsubsection{Noisy sensors}\label{sec:noisySense}

To see how our formalism works, suppose the agent can perform a noisy
sensing action $\noisysensef$ to sense the functional fluent $f$.
This action is the nondeterministic union of the primitive actions
$\thesensef(x,y)$, which denotes that $x$ is sensed, but the true
value of $f$ is $y$.
As we would expect, the precondition for the primitive action is
\[\Poss(\thesensef(x,y),s) \equiv y = f(s);\]
the action $\thesensef(x,y)$ is possible in situation $s$ exactly if
$f(s) = y$.
We assume that
the likelihood of these actions is otherwise independent of $s$, being
dependent only on the actual value of $f$ (which is equal to $y$ by
the above precondition) and the value sensed:
\[\lhood(\thesensef(x,y),s) = \lhood(\thesensef(x,y)).\]
Note that this assumption is satisfied if we assume that the
likelihood is a linear Gaussian function of $x-y$, as discussed in
Section~\ref{sec:beliefs}.  We take $\noisysensef(x)$ to denote the
complex action of observing $x$ on the sensor; thus,
\[\noisysensef(x) \eqdef \pi{y}.\,\thesensef(x,y).\]

We now want to make sure that the assumptions underlying Bayesian
updating hold. To capture the first assumption, that the world does
not change while the observation is being made, we simply ensure that
$\thesensef(x,y)$ does not affect any other fluent besides $\K$ and
$\pr$. In particular, using (\ref{poseffect}) and~(\ref{negeffect}),
we ensure that the action does not appear in the positive or negative
effect axioms for any other fluent.
That the second assumption holds is almost immediate: the value $x$
returned by the sensing action is influenced only by the true value of
$f$.

Now we derive a characterization of the agent's beliefs after
reading the value $x$ from the sensor.
Suppose that $\succes{s}$ is a possible successor state after
reading the sensor, i.e.,
$\Do(\noisysensef(x),s,\succes{s})$ holds, so that $\succes{s}$ is the
result of executing
$\thesensef(x,y)$ in situation $s$, for some $y$ and a fixed value of
$x$. {F}rom the precondition of $\thesensef(x,y)$, we see that we must
have $y=f(s)$, so there is actually only one possible successor state;
that is, $\succes{s} = do(\thesensef(x,f(s)),s)$. Since $\thesensef$
does not affect $f$, we also have $f(\succes{s}) = f(s)$.

{F}rom $\pr$'s successor-state axiom (Defn.~\ref{def:sspr}), we have
that
$\pr(\succes{s'},\succes{s})$ can be nonzero only for situations
$\succes{s'}$ that are the result of executing an observationally
indistinguishable action whose preconditions are satisfied in a state $s'$.
{F}rom the axioms for $\thesensef$, we see that all such situations with
non-zero weight must be of the form $\succes{s'} =
\dofn(\thesensef(x,f(s')),s')$, so $\pr(\succes{s'},\succes{s}) =
\pr(s',s) \lhood(\thesensef(x,f(s')),s')$ for some $s'$, and
$f(\succes{s'})
= f(s')$.

Since $\bel$ is an abbreviation for summations over $\pr$
(Defn.~\ref{def:bel}), we obtain
\[
\begin{array}{l}
  \Do(\noisysensef(x),s,\succes{s}) \supset \\
  \qquad \bel(f(\snow)=t,\succes{s}) \\
  \qquad \mbox{} =
  \left. \sum_{\{\succes{s'}:f(\succes{s'})=
t\}}\pr(\succes{s'},\succes{s})
    \right/ \sum_{\succes{s'}} \pr(\succes{s'},\succes{s}) \\
  \qquad \mbox{} =
  \left. \sum_{\{s': f(s') = t\}}\pr(s',s)
\lhood(\thesensef(x,f(s')),s')
    \right/ \sum_{s'} \pr(s',s)\lhood(\thesensef(x,f(s')),s').
\end{array}
\]
It is not hard to show that the numerator of this last expression is the
(unnormalized) belief that the reading is $t$ and $x$ is observed (which
is the product of the prior belief that $t$ will be read  and the
conditional probability of observing $x$ given $t$), while the
denominator is the (unnormalized) belief that $x$ will be read.
Thus, this computation simulates the Bayesian computation.
This comes out even more clearly if we take advantage of our assumption
that the likelihood is independent of $s$.
With this assumption, the term $\lhood(\thesensef(x,f(s')),s')$ is
constant over the summation in the numerator.  By dividing both
numerator and denominator by $\sum_{s'}\pr(s',s)$, we can
convert the
sums over $\pr$ into
an expression involving
$\bel$:
\begin{equation}
  \label{eq:sensor.update}
  \begin{array}{l}
    \Do(\noisysensef(x),s,\succes{s}) \supset \mbox{} \\
    \qquad \bel(f(\snow)=t,\succes{s}) =
    \frac{\displaystyle \bel(f(\snow) = t,s)\lhood(\thesensef(x,t))}
    {\displaystyle {\sum}_{t'} \bel(f(\snow) =
t',s)\lhood(\thesensef(x,t')).}
  \end{array}
\end{equation}

So we see that the belief that
the value of $f$ is $t$ after doing a sensing action is
the product of the
belief that the value of $f$ was $t$ before doing the sensing action and
the conditional probability
of reading the value $x$ given that the true value is $t$, divided by a
normalizing factor.
Furthermore, if we examine the denominator, we see that
it can be interpreted as the agent's prior belief that
it would read the value $x$ from its sensor.
Again, our calculation is identical to the standard Bayesian calculation.

\subsubsection{Noisy effectors}
\label{sec:noisyEff}
Noisy effectors modify the agent's beliefs in a similar manner.  The
key difference here is that these actions also change the world, so
the agent's updated beliefs must reflect those changes.

Suppose that we have an action $\thef(x,y)$ that can be used to modify
the current value of the fluent $f$. Let its precondition be
\[\Poss(\thef(x,y),s) \equiv \true\]
(indicating that the action is always executable%
\footnote{This is similar
to (\ref{eq:preadvance}), the precondition to $\theAdvance$,
except that we are not asserting any
  absolute error bound.}),
its likelihood function be
\[\lhood(\thef(x,y),s) = \lhood(\thef(x,y))\]
(indicating that the likelihood is not dependent on the situation),
and its corresponding nondeterministic action (denoting that the
effector was actuated with the nominal value $x$) be
\[\noisyf(x) \eqdef \pi{y}.\,\thef(x,y).\]

Furthermore, suppose that the only effect of $\thef$ is to modify the
fluent $f$ (and suppose that it is the only action that does so). This
is reflected in $f$'s successor-state axiom, which
is
\[
\begin{array}{l}
  \Poss(a,s) \supset f(\dofn(a,s)) = \mbox{} \\
  \qquad \mbox{\bf if\ } \exists{x,y}.\,a=\thef(x,y)\\
  \qquad\qquad \mbox{\bf then\ } f(s)+y\\
  \qquad\qquad \mbox{\bf else\ } f(s).
\end{array}
\]

Suppose that the agent attempts to increment the fluent $f$ by executing
the action $\noisyf(x)$.  From the precondition we can see that there
are many different situations $\succes{s}$ such that
$\Do(\noisyf(x),s,\succes{s})$ (unlike the case above for noisy
sensors, where there was a unique successor situation). Each of these
situations is the result of executing
$\thef(x,y)$, for some $y$, in situation $s$.  That $\thef$ changes the
world
is captured by the fact that for each of these successor situations
$\succes{s} = \dofn(\thef(x,y),s)$, we have $f(\succes{s}) = f(s) + y$.

As in our derivation for noisy sensors, we have
\[
\begin{array}{l}
  \Do(\noisyf(x),s,\succes{s}) \supset \\
 \qquad  \bel(f(\snow)=t,\succes{s}) \\
  \qquad \mbox{} =
  \left. \sum_{\{\succes{s'}:f(\succes{s'})=
t\}}\pr(\succes{s'},\succes{s})
    \right/ \sum_{\succes{s'}} \pr(\succes{s'},\succes{s}) \\
  \qquad \mbox{} =
  \left. \sum_y\sum_{\{\succes{s'}: f(\succes{s'}) = t \land
\succes{s'}
      = \dofnsub(\thefsub(x,y),s')\}}\pr(\succes{s'},\succes{s})
  \right/ \sum_{\succes{s'}} \pr(\succes{s'},\succes{s}) \\
  \qquad \mbox{} =
  \left. \sum_y\sum_{\{s': f(s') = t-y\}} \pr(s',s)\lhood(\thef(x,y),s')
  \right/ \sum_y\sum_{s'} \pr(s',s)\lhood(\thef(x,y),s') \\
  \qquad \mbox{} =
  \left. \sum_y\lhood(\thef(x,y))\sum_{\{s': f(s') = t-y\}} \pr(s',s)
  \right/ \sum_y\lhood(\thef(x,y))\sum_{s'} \pr(s',s). \\
\end{array}
\]
Now dividing top and bottom by $\sum_s' \pr(s',s)$ to convert to
beliefs, and simplifying, using the observation that
$\sum_y\lhood(\thef(x,y))$ is
1, we obtain
\begin{equation}
  \label{eq:effector.update}
  \begin{array}{l}
    \Do(\noisyf(x),s,\succes{s}) \supset \mbox{} \\
    \qquad \bel(f(\snow)=t,\succes{s})=\sum_y
\lhood(\thef(x,y))\bel(f(\snow)=t-y,s).
  \end{array}
\end{equation}

There are two differences between this result and our previous result
for noisy sensors. First, we have a summation over $y$, the possible
values that could have been generated by the action. This summation
arises from the fact that any situation prior to the action could have
given rise to a situation in which $f=t$: all that is required is that
an action with the appropriate effect on $f$ be executed. In the 
noisy-sensor case, the actual value of the fluent is not changed, so we
need to consider only the prior situations in which the fluent has
the same value. Second, it seems that there is no normalizing constant
in this case. However, the normalizing constant is present and equal
to $\sum_{t'}\sum_y \lhood(\thef(x,y)) \bel(f(\snow)=t'-y,s)$. It turns
out
that this sum is always equal to 1.

There is another way of deriving this result. Consider the agent's
beliefs about the value of the fluent $f$ prior to executing $\thef$.
The fluent $f$ can be viewed as being a random variable, with the
probability that it takes on any particular value $t$ equal to
$\bel(f(\snow)=t,s)$. When the action $\noisyf(x)$ is executed, the
value
$y$
that it adds to $f$ is also a random variable with distribution
$\lhood(\thef(x,y))$. Hence, the new value of $f$ after executing
$\noisyf(x)$ is the sum of two independent random variables, the old
value of $f$ and the amount added $y$. A standard result from
probability theory is that the distribution of the sum of two
independent
random variables is the {\em convolution\/} of their distributions. Our
result
above shows that the agent's new beliefs about $f$ are in fact the
convolution of its prior beliefs and the distribution
on the amount by which $f$ can change.

\subsection{Examples}

It is
convenient to be able to talk about what happens to the agent's
beliefs and knowledge after reading some particular sequence of values
{}from its sensors. To do this, we introduce a parameterized version of
$\noisyPos$:
\begin{equation}
  \label{eq:noisyposx}
  \noisyPos(x) \eqdef \pi{y}.\,\thePos(x,y),
\end{equation}
where only the value $y$ is chosen (by the value of $\position$ in
the situation). Intuitively, $\noisyPos(x)$ is the ``action'' where
the agent activates its sensor and then observes that the value $x$
has been returned. This action is not actually executable by the
agent, since it cannot choose to execute a $\noisyPos(3.0)$ over, say,
a $\noisyPos(2.8)$. Nevertheless, we can write
$\Do(\noisyPos(3.0),s,\succes{s})$,
which asserts that the agent has
executed a $\noisyPos$ and read the value $3.0$ from its sensor,
ending up in $\succes{s}$.

\begin{example}
  Suppose that the agent is sensing its position $\position$ using
  $\noisyPos$ actions.
  For the purposes of
  this example, let the action-likelihood axiom for $\thePos$ be
  \[
  \lhood(\thePos(x,y),s) = \hspace*{-2em} \\
  \begin{array}[t]{l}
   \qquad \mbox{\bf if\ } x=y \\
    \qquad\qquad \mbox{\bf then\ } 0.5 \\
    \qquad \mbox{\bf else if\ } |x-y| = 1 \\
    \qquad\qquad \mbox{\bf then\ } 0.25 \\
    \qquad \mbox{\bf else } 0
  \end{array}
  \]
  Here we are assuming that $\position$ and the arguments to $\thePos$
  can take on only integer values (i.e., this is the precision of
  these numbers). The axiom specifies that there is zero probability
  that the sensor will read a $\position$ that is greater than 1 unit
  away from the true $\position$, and that the sensor noise is
  independent of other features of the situation.

  Let the agent's
initial beliefs regarding $\position$ be given by
$\bel(\position(\snow) = t, S_0) = 1/8$, for $t \in \{8,9,11,12\}$,
and $\bel(\position(\snow) = 10,S_0) = 1/2$. Initially, the agent does
not ascribe positive probability to any other possible value for
  $\position$. This distribution of beliefs for the various values of
  $\position$ in $S_0$ is shown in Figure~\ref{fig:example-sense}.

  \begin{figure*}[t]
    \begin{center}
      \leavevmode
      \psfig{figure=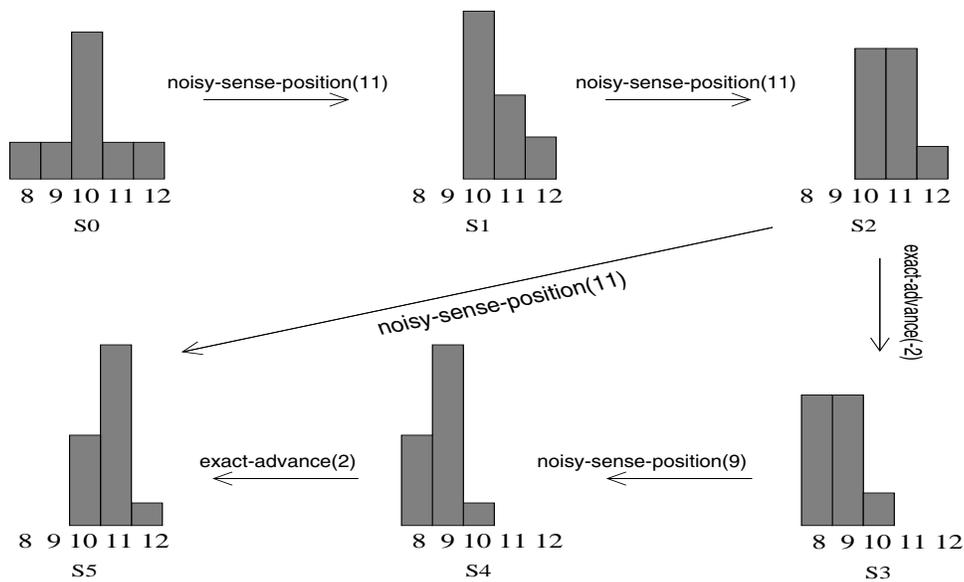,height=3in,width=5in}
    \end{center}
    \caption{Sensing with Exact Motion}
    \label{fig:example-sense}
  \end{figure*}

  Suppose that the agent senses its position and observes the value 11.
If $S_1$ is a possible successor situation, so that
  $\Do(\noisyPos(11),S_0,S_1)$ holds, a simple calculation using
 (\ref{eq:sensor.update}) shows how the agent's beliefs
change:
  $\bel(\position(\snow)=10,S_1) = 4/7$,
$\bel(\position(\snow)=11,S_1) =
  2/7$, and $\bel(\position(\snow)=12,S_1)= 1/7$. The new distribution
is also   shown in Figure~\ref{fig:example-sense}.
Since, with probability 1, the sensor returns a value
that is within 1 unit of the true value,
the agent now has degree of belief zero in the values 8 and 9.

  Note that Figure~\ref{fig:example-sense} shows that the agent still
that the most likely value of $\position$ is 10, even though
  its sensor returned the value 11. This arises {}from the agent's
  high prior belief in
the value being $10$.

  We have not specified an absolute error bound for $\thePos$, i.e., a
  value for $c$ in (\ref{eq:prepos}), implicitly assuming it to be
  greater than 1. If, for example, we were to specify that the agent
  {\em knows\/} that the sensor cannot return an error of greater than
    2, i.e., if we set $c=2$ in the precondition axiom, then we will
    have that the agent knows that $\position\neq 8$. However, the
    agent will not know that $\position\neq 9$, even though it still
    has zero degree of belief in this value.
Our framework distinguishes
    between full belief and knowledge
in this case.
\end{example}

Sequences of sensor readings of the same fluent, including sequences
of readings {}from different sensors, are also handled correctly in
our framework. Such sequences correspond to sequences of sensing
actions, and thus are handled by a simple iteration of
(\ref{eq:sensor.update}), given that the sensors satisfy the
assumptions of that equation. In particular, if the sensor action
likelihood function is dependent only on the actual value being
sensed, then each sensor reading will be independent of all previous
readings.%
\footnote{A situation is a world history consisting of the sequence of
  actions executed (Section~\ref{sec:sitcalc}). Hence, one can easily
  write a sensor-likelihood axiom that did display a dependency on
  previous readings, if, for example, one wanted to capture the
  properties of a sensor that displayed some form of hysteresis.} %
As a result, after a sequence of sensing actions, the agent will come
to have either greater or less certainty about the value of the sensed
fluent, depending on whether or not the sequence of readings agree or
not.

\begin{example}
  Suppose that the agent executes another $\noisyPos$ action in the
  situation $S_1$. Further, suppose that the agent observes the same
  value as before 11, and let $S_2$ be such that
  $\Do(\noisyPos(11),S_1,S_2)$. Then, another application of
(\ref{eq:sensor.update}) (applied to the agent's beliefs in
  $S_1$), yields the belief distribution shown in
  Figure~\ref{fig:example-sense}: $\bel(\position(\snow)=10,S_2) =
4/9$,
  $\bel(\position(\snow)=11,S_2) = 4/9$, and
$\bel(\position(\snow)=12,S_2)=
1/9$.
  The agent's beliefs have converged more tightly around the value 11,
  since it has now sensed that value twice.
\end{example}

The most important part of our formalism is the natural manner in
which effectors and sensors interact in their effects on the agent's
beliefs. For simplicity, first we examine the case where the agent has
exact control over a fluent, but can sense its value only
approximately. Then we examine the case of a noisy sensor
interleaved with a noisy effector.

\begin{example}
  Suppose that instead of the noisy effector $\noisyAdvance$, the agent
  has an exact effector action $\exactAdvance$ that has no
  preconditions, and affects $\position$ as specified in
 (\ref{eq:ssexactpos}). Note that $\exactAdvance$ is an ordinary
  deterministic action, so it is observationally indistinguishable only
  {}from itself, and it has a unit likelihood function:
  $\lhood(\exactAdvance(x),s) = 1$.

  Let the agent move exactly 2 units backwards when it is in situation
  $S_2$. Call the new situation $S_3 = \dofn(\exactAdvance(-2),S_2)$.
  Then, the successor-state axiom for $\pr$ and $\position$ imply that
  the agent's beliefs are precisely shifted to worlds in which it has
  moved backward 2 units: we have a simple transfer of probability mass
  from each situation to its successor situation in which the agent
  has moved backward 2 units.

  Deterministic actions like this modify the agent's beliefs in a
  manner that is related to Lewis's notion of {\em imaging\/}
  \cite{Lewis76}. In imaging, beliefs are updated by transferring all
  probability mass to the ``closest'' world, rather than by
  renormalizing the mass after removing some worlds as when we
  condition. Here, every situation transfers its probability mass to
  its successor. Different things are true in these successor
  situations, as the action has effected various changes. This means
  that the agent will believe different things in the successor
  situation. Nevertheless, since all of these successor situations
  arise from the execution of the same action, the changes to the
  agent's beliefs are generally systematic. In particular, in the
  absence of ramifications, it will change its beliefs only about
  fluents affected by the action.

  In this example, we have that $\bel(\position(\snow) = t,
S_3)
=
  \bel(\position(\snow) = t+2, S_2)$ for all $t$. The agent's shifted
beliefs are shown
  in Figure~\ref{fig:example-sense}. This is exactly how one would
  expect the agent's beliefs to change in response to an exact
  movement like this.
\end{example}

\begin{example}
  Suppose that the agent again executes a $\noisyPos$ action in $S_3$
and observes the value 9. Let $S_4$ be such that
  $\Do(\noisyPos(9),S_3,S_4)$. A reading of 9 is consistent with the
  previous readings of 11, since the agent has moved back 2 units.
  Hence, as shown in Figure~\ref{fig:example-sense}, it results in a
  further tightening of the agent's beliefs, around the value 9. If
  the agent subsequently moves forward 2 units, executing an
  $\exactAdvance(2)$ action, so that $S_5 = \dofn(\exactAdvance(2),
  S_4)$, its beliefs will then be clustered around 11, as shown in the
  figure: $\bel(\position(\snow)=10,S_5) = 4/13$,
$\bel(\position(\snow)=11,S_5)
=
  8/13)$, and $\bel(\position(\snow)=12,S_5)= 1/13$.

  Intuitively, since the agent's $\exactAdvance$ action incurs no
  error, we would expect that if the agent had sensed the value 11 in
  situation $S_2$, then its beliefs about the distance to the wall
  should not change after moving forwards and backwards an equal
  distance. Our model respects this intuition. In particular, the
  agent's beliefs in $S_5$ are identical to what they would be if it
  had executed an $\noisyPos(11)$ in $S_2$, as indicated in the figure
  by the diagonal arrow {}from $S_2$ to $S_5$.
\end{example}

\begin{example}
  Now we examine the case of a noisy effector. Let us return to the
  effector $\noisyAdvance$, taking (\ref{eq:sspos}) to specify the
successor-state
  axioms for $\position$. Suppose that the likelihood axiom for
  $\theAdvance$ is
  \[
  \lhood(\theAdvance(x,y),s) = \hspace*{-2em}
  \begin{array}[t]{l}
    \qquad \mbox{\bf if\ } x=y \\
    \qquad\qquad \mbox{\bf then\ } 0.5 \\
    \qquad \mbox{\bf else-if\ } |x-y| = 1 \\
    \qquad\qquad \mbox{\bf then\ } 0.25 \\
    \qquad \mbox{\bf else } 0
  \end{array}
  \]
Here, with probability 1, the difference between the agent's actual move
and the move specified is no more than one unit.

  Starting at situation $S_5$, with beliefs about $\position$ as given
  above, suppose that $S_6$ is a successor state of an attempt to move
  forward 1 unit, so that $\Do(\noisyAdvance(1),S_5,S_6)$ holds. The
  likelihood axiom above indicates that in this case the probability the
  agent moves forward 2 units is $1/4$, 1 unit is $1/2$, and 0 units
  is $1/4$. Equation~\ref{eq:effector.update} can be used to calculate
  the agent's new beliefs about its position. In particular, we obtain
  $\bel(\position(\snow)=10,S_6) = 4/52$,
$\bel(\position(\snow)=11,S_6) = 16/52$,
  $\bel(\position(\snow)=12,S_6)= 21/52$,
$\bel(\position(\snow)=13,S_6)=
10/52$,
  and $\bel(\position(\snow)=14,S_6)= 1/52$. This distribution is shown
in
  Figure~\ref{fig:example-move}.

  \begin{figure*}[t]
    \begin{center}
      \leavevmode
      \psfig{figure={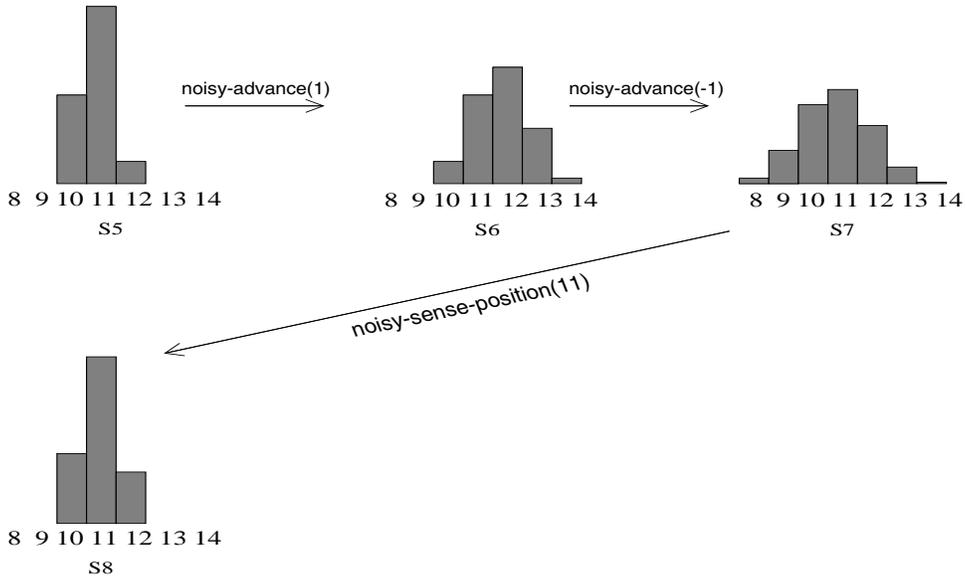},height=3in,width=5in}
    \end{center}
    \caption{Noisy Motion}
    \label{fig:example-move}
  \end{figure*}

  Suppose that the agent now attempts to move back to its previous
  position by executing a $\noisyAdvance(-1)$. If situation $S_7$ is
  such that $\Do(\noisyAdvance(-1),S_6,S_7)$ holds, then the agent's
beliefs in $S_7$ are
  $\bel(\position(\snow)=8,S_7) = 1/52$, $\bel(\position(\snow)=9,S_7)
=
6/52$,
  $\bel(\position(\snow)=10,S_7) = 14.25/52$,
$\bel(\position(\snow)=11,S_7)
=
  17/52$, $\bel(\position(\snow)=12,S_7)= 10.5/52$,
$\bel(\position(\snow)=13,S_7)=
  3/52$, and $\bel(\position(\snow)=14,S_7)= 0.25/52$.

  We see from this example that in the presence of noisy motion, the
  agent's confidence in its location degrades; its beliefs spread out.
  And this effect increases the more it moves. Unless it wants to lose
  its way completely, it must resort to sensing.
  Figure~\ref{fig:example-move} shows what happens to the agent's
  beliefs after it senses its position to be 11 in situation
  $S_7$. If $S_8$ is such that
  $\Do(\noisyPos(11),S_7,S_8)$ holds, then the agent's new beliefs in
$S_8$ are
  $\bel(\position(\snow)=8,S_8) = 0$, $\bel(\position(\snow)=9,S_8) =
0$,
  $\bel(\position(\snow)=10,S_8) = 57/235$,
$\bel(\position(\snow)=11,S_8)
=
136/235$,
  $\bel(\position(\snow)=12,S_8)= 42/235$,
$\bel(\position(\snow)=13,S_8)=
0$,
  and $\bel(\position(\snow)=14,S_8)= 0$.  Thus, the beliefs are
considerably sharpened by the
  sensing operation.
\end{example}

The previous examples have employed discrete beliefs and likelihood
functions, which make the computations more apparent. More practical
models are also easily accommodated, and in fact sometimes yield
simpler, albeit less obvious, computations.

In robotics applications one of the most popular models of noisy
sensors and effectors is the linear Gaussian model where the nominal
value is the actual value plus a Gaussian noise factor (see, e.g., the
various papers in \cite{RUR}). Equations~\ref{eq:lkadvance}
and~\ref{eq:lkpos} are examples of linear Gaussian models. The next
example shows how easy belief update is with such models.

\begin{example}
  Suppose that the agent has an action $\noisyf$ for effecting changes to
  a fluent $f$, and an action $\noisysensef$ for sensing its value, as
  in Sections~\ref{sec:noisySense} and~\ref{sec:noisyEff}. Further,
  suppose that the likelihood functions for both $\thef$ and
$\thesensef$ are linear Gaussian, so that
  \[\lhood(\thef(x,y),s) = \norm((y-x)/\sigma_e),\]
  and
  \[
   \lhood(\thesensef(x,y),s) = \norm((y-x)/\sigma_s).\]

  Suppose that the agent's initial beliefs
  about a fluent $f$
  are described by the normal distribution $N(\mu_0,\sigma_0^2)$ with
  mean $\mu$ and variance $\sigma_0^2$. That is, $\bel(f(\snow)=t,S_0)
=
  \norm((t-\mu_0)/\sigma_0)$, where $\norm(z)$ is the standardized
  normal distribution with mean 0 and variance 1.

  Now suppose that a $\noisyf(x_1)$ action is executed.
  Equation~\ref{eq:effector.update} shows that the agent's new beliefs
  about $f$ is that its value is the result of adding two independent
  random variables (which gives rise to the convolution of two
  distributions). In particular, the new value for the fluent $f$ is
  the sum of the change actually generated by the action
  and $f$'s previous value. The likelihood axiom for $\thef(x_1,y)$
  indicates that $y$, the change to $f$ generated by the action, is a
  normally distributed random variable with mean $x_1$ and variance
  $\sigma_e^2$. The agent's initial beliefs indicate that $f$'s
  previous value is also a normally distributed random variable with
  mean $\mu_0$ and variance $\sigma_0^2$.

  A well known (and easy-to-derive) result is that the sum of two
  normally distributed random variables is also normally distributed
  (e.g., see \cite[page 146]{Lindley-I}). In particular, we have that,
after executing the action, the agent's new beliefs about $f$ are
  described by another normal distribution with mean $\mu_1 =
  \mu_0+x_1$ and variance $\sigma_1^2=\sigma_0^2+\sigma_e^2$.

  Next, suppose that the agent executes a sensing action and reads the
  value $z_1$ from its sensor, i.e., the action $\noisysensef(z_1)$ is
  executed. Equation~\ref{eq:sensor.update} indicates that the agent's
  new beliefs about the value of $f$ are the normalized result of
  multiplying its prior beliefs by the probability of observing $z_1$
  given the value of $f$. Both of these quantities are again described
  by normal distributions. The prior beliefs are normally distributed,
  and for each value of $f$, the probability of $z_1$ is also normally
  distributed, according to the likelihood function above. In this case,
another
  well-known result is that the posterior beliefs about $f$ remain
  normally distributed (see, e.g., \cite[page 2]{Lindley-II}). In
  particular, we have that the agent's new beliefs about $f$ are that
  its value is normally distributed with mean $\mu_2 = (z_1\sigma_1^2
  + \mu_1\sigma_s^2)/(\sigma_s^2+\sigma_1^s)$ and variance $\sigma_2^2
  = \sigma_1^2\sigma_s^2/(\sigma_1^2 + \sigma_s^2)$.

  It is easy to see that these equations can be applied iteratively to
  keep track of the agent's beliefs about a fluent's value given a
  sequence of modification to and sensing of this value. With a normal
  distribution we need keep track only of the current mean and
  variance to completely describe the agent's beliefs.

  The fact that a normal distribution is preserved under these types
  of updates, with easy-to-compute modifications to its parameters,
  forms the basis of the popular technique of Kalman filtering
  \cite{Dean-Wellman}. In fact, this example is simply an instance of
  Kalman filtering, and it can be generalized within our framework to
  handle the situation where the fluent being modified and sensed is
  vector valued. In this case the above variances would be replaced by
  covariance matrices, and the computations to update the mean and
  variance would become matrix manipulations.
\end{example}

All of the previous examples involved the sensing or affecting of a
numeric fluent. Our final example demonstrates that the formalism can
also be used to model noisy actions that affect non-numeric fluents.

\begin{example}
  In Section~\ref{sec:sitcalc} we used the example of dropping fragile
  objects. However, when a fragile object is dropped it does not
  always break, it breaks with some probability. We an model this
  situation quite easily in our formalism.

  Suppose that there are two primitive drop actions, $\dropbreak$ and
$\dropnobreak$,
  characterized by the following axioms:
  \[\Poss(\dropbreak(x),s) \equiv \holding(x,s) \land
  \fragile(x,s)\]
  \[\Poss(\dropnobreak(x),s) \equiv \holding(x,s)\]
  \[\ObsIn(\dropbreak(x),a',s) \equiv a' = \dropbreak(x) \lor a' =
  \dropnobreak(x)\]
  \[\ObsIn(\dropnobreak(x),a',s) \equiv a' = \dropnobreak(x) \lor a' =
  \dropbreak(x)\]
  \[
  \lhood(\dropbreak(x),s) =
  \mbox{\bf if\ } \fragile(x)
  \quad \mbox{\bf then\ } 0.8
  \quad \mbox{\bf else\ } 0
  \]
  \[
  \lhood(\dropnobreak(x),s) =
    \mbox{\bf if\ } \fragile(x)
    \quad \mbox{\bf then\ } 0.2
    \quad \mbox{\bf else\ } 1.
  \]
  Let the successor-state axiom for $\broken$ be
  \[
    \Poss(a,s) \supset
    \broken(x,\dofn(a,s)) \equiv  a = \dropbreak(x) \lor \broken(x,s).
  \]
  Finally, suppose that the agent can execute a nondeterministic action
  $\drop$:
  \[\drop(x) \eqdef \dropbreak(x) | \dropnobreak(x). \]

  These axioms specify that  the actions $\dropbreak$ and
  $\dropnobreak$  cannot be distinguished by the agent. That is, the
  agent can execute a $\drop$ action, but does not know ahead of time
  whether or not dropping an object will cause it to be broken. If the
  object is not fragile only the $\dropnobreak$ action can be
  executed, in which case it has probability 1 of being executed.
  On the other hand, if the object is fragile, then the preconditions of
  both $\dropnobreak$ and $\dropbreak$ are satisfied and either can be
  executed. The probability of $\dropbreak$ being executed, i.e., the
  probability that a fragile object will break when dropped, is 0.8,
  while $\dropnobreak$ has probability 0.2. The successor-state axiom
  for break specifies that $\dropbreak$ does in fact cause an object
  to become broken.

  If the agent executes $\drop(x)$ in situation $s$ for some object
  $x$ given that $\Know(holding(x),s)$ (so that the action is
  possible), then we obtain for any successor situation $\succes{s}$
  \[
  \begin{array}{l}
    \Do(\drop(A),s,\succes{s}) \land \Know(holding(x),s)\\
    \qquad \mbox{} \supset
    \bel(\broken(x,\snow),\succes{s})= 0.8 \times
\bel(\fragile(x,\snow),s).
  \end{array}
  \]
  For example, if the agent's degree of belief is 0.5 that the object
  being dropped is fragile, then the agent will
have degree of belief degree 0.4
  that the object will be broken after being dropped. Similarly, if
  the agent is certain that the object is fragile, then it will have
  degree of belief 0.8
that the object will break after being dropped.
  This makes sense intuitively, as the object can be broken only by
  dropping if it was originally fragile, and even then there is a
  probability of only 0.8 that it will be broken by the drop action.
\end{example}

\section{Conclusion}\label{sec:conclusion}
We have demonstrated that noisy perception and actions can be modeled
in the situation calculus by a simple extension of previous work. In
particular, from the successor-state axiom for $\pr$,
(\ref{def:sspr}), and a constraint on its values in $S_0$,
(\ref{eq:prob}), we obtain as consequences what many have argued to
be the natural models for belief update from perception (Bayesian
conditioning) and from actions (a form of Lewis's imaging)
\cite{Pearl:SSS95}.  Most importantly, our formalism succeeds in
capturing some key features of the interaction between these two
models for belief change.

Much of our approach can be exported to alternate formalisms.  For
example, instead of the situation calculus a modal logic could have
been used. Similarly, the probabilistic component could be replaced
with an alternate formalism, like Dempster-Shafer belief functions
\cite{Shaf} or possibility measures \cite{DP88}. All that would be
required is to replace the functional fluent $\pr$ and axioms for
$\bel$ with fluents and axioms to support an alternate measure of
belief. The likelihood functions could then be replaced with
non-probabilistic functions to support an alternate rule of belief
update.

As for future work, apart {}from addressing limitations of the
formalism, there is its application to high-level agent control. In
the {\sc golog} work mentioned in the introduction, the ability of an
agent to execute a program depends on what it {\em knows\/} about the
truth value of the test conditions in that program
\cite{lesperance:ability}.  When an agent has only a degree of belief
in the truth of a test condition in a program
it can randomize the program's execution. That is, instead of
executing a set of actions only when a test condition is true, it can
randomly decide to execute the actions with probability $p$ when its
degree of belief in the test condition is $p$. The key issue will then
be characterizing the program's effects. For example, one could
construct programs that with probability bounded below by $1-\epsilon$
achieve a certain condition. Or one could employ a richer model that
included an assignment of utilities to situations, and attempt to
construct programs that maximize the agent's expected utility.

\appendix
\section{Formalization of Belief}
\label{app:belief}

In this appendix, we show how $\bel$ can be formalized so as to ensure
that it is in fact a probability distribution.

As mentioned in Section~\ref{sec:sitcalc}, we treat the situation
calculus as a many-sorted dialect of the predicate calculus with some
second-order features. Included among the sorts are situations and
real numbers. For convenience, we also assume that the language
includes the natural numbers as a sub-sort of the reals.%
\footnote{As is well known, with 0, 1, and $+$ in the language, we can
actually define the natural numbers, since we have second-order
quantification, as follows: $n \in N$ is an abbrevation for $\forall
P((0 \in P \land \forall y (y \in P \supset y+1 \in P)) \supset n \in
P)$.}  
In
particular, in the language we have variables that range only over
these specific sorts.\footnote{Note that we can easily do without such
  variables by including some unary ``type'' predicates in our
  language.}

With sorted variables we can then formalize the summations used in our
definition of $\bel$ (Defn.~\ref{def:bel}). Let $\phi(\snow)$ be a
formula over the special situation term $\snow$, $r$ and $r'$ be
variables of sort real, $f$ and $g$ be second-order function variables
(ranging over all functions), and $m$, $i$, and $j$ be variables of
sort natural number. We define $\sump(\phi(\snow),s)$ to be the
summation of $\pr$ in $s$ over all situations $s'$ satisfying
$\phi(\snow/s')$:
\[
\begin{array}{l}
  \sump(\phi(\snow),s) = r 
\eqdef \\
  \qquad\mbox{} \forall r'. (r' < r) \equiv \\
  \qquad\qquad\mbox{} \exists f,g,m. \\
  \qquad\qquad\qquad\mbox{}
  \forall i,j. (i\neq j \supset g(i) \neq g(j)) \land
               (i \leq m \supset \phi(\snow/g(i))) \\
  \qquad\qquad\qquad\mbox{} \land  f(0) = 0 \\
  \qquad\qquad\qquad\mbox{} \land
         \forall i. f(i+1) = f(i) + \pr(g(i),s)  \\
  \qquad\qquad\qquad\mbox{} \land  f(m) > r'
\end{array}
\]
Basically, this formula says that $\sump(\phi(\snow),s)$ is equal to
$r$ iff for every value $r'$, $r'$ is less than $r$ iff there exists a
finite set $m$ of situations (enumerated by the function $g(0)$,
\ldots, $g(m)$) satisfying $\phi$ whose $\pr$ values sum 
to a value (computed by the function $f$) greater than $r'$. Note that
this definition entails that an infinite set of situations has a $\pr$
sum 
that is the limit of the sum of $\pr$ over its elements. In
particular, the resulting probability distribution (achieved once we
normalize the sum) will be a discrete distribution.

Now we can formally define $\bel(\phi(\snow),s)$ as
\[
\bel(\phi(\snow),s) = \sump(\phi(\snow),s)/\sump(\true,s),
\]
where $\true$ is satisfied by all situations. 

In Section~\ref{sec:beliefs} we imposed the following constraints on
$\pr$:
\[\forall s.(\pr(s,S_0) \geq 0) \land (\lnot \K(s,S_0)
\supset\pr(s,S_0) = 0).\] 
Now we can add the additional constraint:
\[\exists r. r>0 \land \sump(\true,S_0) = r,\]
where $r$ is a variable of sort real. This constraint ensures that the
sum of $\pr$ over all situations is, in $S_0$, a finite positive
number. From these constraints it follows, that for every formula
$\phi$, $\sump(\phi(\snow),S_0)$ exists and, since $\pr$ is
non-negative, is less than (or equal to) $\sump(\phi(\snow),S_0)$. It
is then immediate that in $S_0$, $\bel$ is a probability distribution
over the space of situations. Finally, the constraint $\lnot\K(s,S_0)
\supset \pr(s,S_0) = 0$ entails that all that is known is assigned
degree of belief 1.

What about $\bel$ in situations other than $S_0$? By identical
reasoning, in any situation $s$, $\bel$ will be a probability
distribution over the set of situations if the sum of $\pr$ over all
situations is, in $s$, a finite positive number and $\pr$ is
non-negative. To guarantee this we require that the action-likelihood
functions (Defn.~\ref{def:lkaxioms}) be probability distributions over
the set of actions.

First we define $\suml(\phi(\anow,s),s)$ to be the sum of $\lhood(\anow,s)$
over all actions $\anow$ that satisfy the formula $\phi(\anow,s)$ in situation
$s$. This can be defined in an identical manner to $\sump$, the only
change required is to sum the values of $\lhood(g(i),s)$ instead of
the values of $\pr(g(i),s)$ in the formula above. We then impose the
following constraints:
\begin{enumerate}
\item $\forall a,s. \lhood(a,s) \geq 0$.
\item $\forall a,s.\suml(\ObsIn(a,\anow,s),s) = 1$.
\end{enumerate}
The first formula says the action-likelihood function is non-negative,
while the second formula ensures that for every action $a$ the sum of
the likelihood function over all actions $\anow$ that are observationally
indistinguishable to $a$ in situation $s$ is one.  Together these two
formulas make the action-likelihood function a discrete probability
distribution over each set of observationally indistinguishable
actions.

With this in hand, the successor-state axiom for $\pr$
(Defn.~\ref{def:sspr}) gives us what we require. Given that
$\sump(\true,s)$ is finite the axiom shows that for any action $a$,
$\sump(\true,\dofn(a,s))$ remains finite since in the term $\pr(s',s)
\times \lhood(a',s')$, $\lhood(a',s')$ has a finite sum over the $a'$.
Since $\sump(\true,S_0)$ is finite, we obtain by induction that it
remains finite for every situation (since all situations arise from
applying a finite sequence of actions to $S_0$). Finally, the
successor-state axiom for $\K$ (Defn.~\ref{def:ssK}) shows that only
$K$-related situations get positive values of $\pr$. Hence in every
situation, not just $S_0$, all that is known is assigned degree of belief
1.

\bibliographystyle{plain}
\bibliography{bhl}

\end{document}